\documentclass[aos,MSNbibl,nameyear,seceqn,dvips]{arximspdf}
\usepackage{mathrsfs}
\usepackage{dcolumn}
\usepackage{graphicx}

\doi{10.1214/11-AOS887}
\volume{39}
\issue{4}
\pubyear{2011}
\firstpage{1878}
\lastpage{1915}

\makeatletter

\newcolumntype{d}[1]{D{.}{.}{#1}}

\newcommand{\popx}{{\mathscr{X}}}
\newcommand{\popw}{{\mathscr{W}}}
\newcommand{\Wn}{{W^{(n)}}}
\newcommand{\popwn}{{\mathscr{W}^{(n)}}}
\newcommand{\popd}{{\mathscr{D}}}
\newcommand{\popdni}{{\mathscr{D}_{ii}^{(n)}}}
\newcommand{\popl}{{\mathscr{L}}}
\newcommand{\Ln}{{L^{(n)}}}
\newcommand{\popln}{{\mathscr{L}^{(n)}}}
\newcommand{\R}{\mathcal{R}}

\newcommand{\bP}{\mathbb{P}}
\newcommand{\E}{\mathbb{E}}

\newtheorem{theorem}{Theorem}[section]
\newproclaim{definition}{Definition}
\newproclaim{Example}{Example}[section]

\newtheorem{lemma}{Lemma}[section]
\newtheorem{prop}{Proposition}[section]

\makeatother

\begin{document}
\begin{frontmatter}

\title{Spectral clustering and the high-dimensional stochastic blockmodel\thanksref{T1}}
\runtitle{Clustering for the stochastic blockmodel}

\thankstext{T1}{Supported by Army Research Office
Grant ARO-W911NF-11-1-0114; an NSF VIGRE Graduate Fellowship; NSF
Grants CCF-0939370, DMS-07-07054, DMS-09-07632 and SES-0835531 (CDI);
and a Sloan Research Fellowship.}

\begin{aug}
\author[A]{\fnms{Karl} \snm{Rohe}\corref{}\ead[label=e1]{karlrohe@stat.berkeley.edu}},
\author[A]{\fnms{Sourav} \snm{Chatterjee}\ead[label=e2]{sourav@stat.berkeley.edu}} and
\author[A]{\fnms{Bin} \snm{Yu}\ead[label=e3]{binyu@stat.berkeley.edu}}
\runauthor{K. Rohe, S. Chatterjee and B. Yu}
\affiliation{University of California, Berkeley}
\address[A]{Department of Statistics\\
University of California\\
Berkeley, California 94720\\
USA\\
\printead{e1}\\
\hphantom{E-mail: }\printead*{e2}\\
\hphantom{E-mail: }\printead*{e3}} 
\end{aug}

\received{\smonth{7} \syear{2010}}
\revised{\smonth{12} \syear{2010}}

%
\begin{abstract}
Networks or graphs can easily represent a diverse set of data sources
that are characterized by interacting units or actors. Social networks,
representing people who communicate with each other, are one example.
Communities or clusters of highly connected actors form an essential
feature in the structure of several empirical networks. Spectral
clustering is a popular and computationally feasible method to discover
these communities.

The stochastic blockmodel [\textit{Social Networks} \textbf{5} (1983)
109--137] is a social network model with well-defined communities; each
node is a member of one community. For a network generated from the
Stochastic Blockmodel, we bound the number of nodes ``misclustered'' by
spectral clustering. The asymptotic results in this paper are the first
clustering results that allow the number of clusters in the model to
grow with the number of nodes, hence the name high-dimensional.

In order to study spectral clustering under the stochastic blockmodel,
we first show that under the more general latent space model, the
eigenvectors of the normalized graph Laplacian asymptotically converge
to the eigenvectors of a ``population'' normalized graph Laplacian.
Aside from the implication for spectral clustering, this provides
insight into a graph visualization technique. Our method of studying
the eigenvectors of random matrices is original.
\end{abstract}

%

\end{frontmatter}

\section{Introduction}\label{sec1}

Researchers in many fields and businesses in several industries have
exploited the recent advances in information technology to produce an
explosion of data on complex systems. Several of the complex systems
have interacting units or actors that networks or graphs can easily
represent, providing a range of disciplines with a suite of potential
questions on how to produce knowledge from network data. Understanding
the system of relationships between people can aid both epidemiologists
and sociologists. In biology, the predator-prey pursuits in a natural
environment can be represented by a food web,
helping researchers
better understand an ecosystem. The chemical reactions between
metabolites and enzymes in an organism can be portrayed in a metabolic
network, providing biochemists with a tool to study metabolism.
Networks or graphs conveniently describe these relationships,
necessitating the development of statistically sound methodologies for
exploring, modeling and interpreting networks.\looseness=1

Communities or clusters of highly connected actors form an essential
feature in the structure of several empirical networks. The
identification of these clusters helps answer vital questions in a
variety of fields. In the communication network of terrorists, a
cluster could be a terrorist cell; web pages that provide hyperlinks to
each other form a community that might host discussions of a similar
topic; and a community or cluster in a social network likely shares a
similar interest.

Searching for clusters is algorithmically difficult because it is
computationally intractable to search over all possible clusterings.
Even on a relatively small graph, one with 100 nodes, the number of
different partitions exceeds some estimates of the number of atoms in
the universe by twenty orders of magnitude [\citet{atomsinuniverse}].
For several different applications, physicists, computer scientists
and statisticians have produced numerous algorithms to overcome these
computational challenges. Often these algorithms aim to discover
clusters which are approximately the ``best'' clusters as measured by
some empirical objective function [see \citet
{fortunato2009community} or
\citet{fjallstrom1998algorithms} for comprehensive reviews of these
algorithms from the physics or the engineering perspective,
resp.].

Clustering algorithms generally come from two sources: from fitting
procedures for various statistical models that have well-defined
communities and, more commonly, from heuristics or insights on what
network communities should look like. This division is analogous to the
difference in multivariate data analysis between parametric clustering
algorithms, such as an EM algorithm fitting a Gaussian mixture
model, and nonparametric clustering algorithms such as $k$-means, which
are instead motivated by optimizing an objective function.
\citet{snijders1997estimation}, \citet{nowicki2001estimation},
\citet{handcock2007model} and
\citet{airoldi2008mms} all attempt to cluster the nodes of a
network by
fitting various network models that have well-defined communities. In
contrast, the Girvan--Newman algorithm [\citet
{girvan2002community}] and spectral clustering are two algorithms in a
large class of algorithms motivated by insights and heuristics on
communities in networks.

\citet{newman2004finding} motivate their algorithm by observing, ``If
two communities are joined by only a few inter-community edges, then
all paths through the network from vertices in one community to
vertices in the other must pass along one of those few edges.'' The
Girvan--Newman algorithm searches for these few edges and removes them,
resulting in a graph with multiple connected components (connected
components are clusters of nodes such that there are no connections
between the clusters).\vadjust{\goodbreak} The Girvan--Newman algorithm
then returns these
connected components as the clusters. Like the Girvan--Newman
algorithm, spectral clustering is a ``nonparametric'' algorithm
motivated by the following insights and heuristics: spectral clustering
is a convex relaxation of the normalized cut optimization problem
[\citet{shi2000normalized}], it can identify the connected
components in a graph (if there are any)
[\citet{donath1973lower}, \citet{fiedler1973algebraic}],
and it has an intimate
connection with electrical network theory and random walks on graphs
[\citet{klein1993resistance}, \citet{meila2001random}].

\subsection{Spectral clustering} Spectral clustering is both popular
and computationally feasible [\citet{vonluxburg2007tsc}]. The
algorithm has been rediscovered and reapplied in numerous different
fields since the initial work of \citet{donath1973lower} and
\citet
{fiedler1973algebraic}. Computer scientists have found many
applications for variations of spectral clustering, such as load
balancing and parallel computations [\citet{van1995improved},
\citet{hendrickson1995improved}], partitioning circuits for
very large-scale integration design [\citet{hagen1992new}] and
sparse matrix partitioning [\citet{pothen1990partitioning}].
Detailed histories of spectral clustering can be found in \citet
{spielman2007spectral} and \citet{vonluxburg2008csc}.

The algorithm is defined in terms of a graph $G$, represented by a
vertex set and an edge set. The vertex set $\{v_1, \ldots, v_n\}$
contains vertices or nodes. These are the actors in the systems
discussed above. We will refer to node~$v_i$ as node $i$. We will only
consider unweighted and undirected edges. So, the edge set contains a
pair $(i,j)$ if there is an edge, or relationship, between nodes $i$
and $j$. The edge set can be represented by the adjacency matrix $W \in
\{0,1\}^{n \times n}$:
%
%
\begin{equation} \label{Wdef}
W_{ji} = W_{ij} = \cases{
1, &\quad if $(i,j)$ is in the edge set,\vspace*{2pt}\cr
0, &\quad otherwise.}
\end{equation}
Define $L$ and diagonal matrix $D$ both elements of $\R^{n \times n}$
in the following way:
%
%
\begin{eqnarray}
\label{Ldef1}
D_{ii} & = & \sum_k W_{ik},\nonumber\\[-8pt]\\[-8pt]
L &=& D^{-1/2}WD^{-1/2}.
\nonumber
\end{eqnarray}
Some readers may be more familiar defining $L$ as $I -
D^{-1/2}WD^{-1/2}$. For spectral clustering, the difference is
immaterial because both definitions have the same eigenvectors.

The spectral clustering algorithm addressed in this paper is defined as
follows:\vspace{.2 in}

\begin{center}
\framebox[4.9 in][c]{
\begin{minipage}[l]{4.3in}
\hspace{-.3in} \texttt{Spectral clustering for $k$ many clusters}

\hspace{-.3in}
\texttt{Input: Symmetric adjacency matrix $W \in\{0,1\}^{n\times n}$.\vspace
{.1in}\\
1. Find the eigenvectors \mbox{$X_1, \ldots, X_k \in\R^n$} \mbox
{corresponding to the $k$ eigenvalues of $L$ that}\\
\mbox{are largest in absolute value. $L$ is symmetric,}\\
{so
choose these eigenvectors to be orthogonal.} {Form the matrix $X =
[X_1, \ldots, X_k]
\in\R^{n \times k}$ by putting} {the eigenvectors into the
columns.}\vspace{.1in}\\
2. \mbox{Treating each of the $n$ rows in $X$ as a point} \mbox{in $\R
^k$, run $k$-means
with $k$ clusters. This creates} \mbox{$k$ nonoverlapping sets $A_1,
\dots, A_k$
whose union is} $1, \ldots, n$. \vspace{.1in}\\
Output: $A_1, \ldots, A_k$. This means that node $i$ is assigned to
cluster $g$ if the $i$th row of $X$ is assigned to $A_g$ in step 2.}
\end{minipage}
}
\end{center}

\vspace{.2 in}

Traditionally, spectral clustering takes the eigenvectors of $L$
corresponding to the largest $k$ eigenvalues. The algorithm above takes
the largest $k$ eigenvalues \textit{by absolute value}. The reason for
this is explained in Section \ref{sec3}.

Recently, spectral clustering has also been applied in cases where the
graph $G$ and its adjacency matrix $W$ are not given, but instead
inferred from a measure of pairwise similarity $k(\cdot, \cdot)$
between data points $X_1,\ldots, X_n$ in a~metric space. The similarity
matrix $K \in\R^{n \times n}$, whose $i,j$th element is $K_{ij} =
k(X_i, X_j)$, takes the place of the adjacency matrix $W$ in the above
definition of $L, D,$ and the spectral clustering algorithm. For image
segmentation, \citet{shi2000normalized} suggested spectral
clustering on
an inferred network where the nodes are the pixels and the edges are
determined by some measure of pixel similarity. In this way, spectral
clustering has many similarities with the nonlinear dimension reduction
or manifold learning techniques such as Diffusion maps and Laplacian
eigenmaps [\citet{coifman2005gdt}, \citet{belkin2003led}].

The normalized graph Laplacian $L$ is an essential part of spectral
clustering, Diffusion maps and Laplacian eigenmaps. As such, its
properties have been well studied under the model that the data points
are randomly sampled from a probability distribution, whose support may
be a manifold, and the Laplacian is built from the inferred graph based
on some measure of similarity between data points.
\citet{belkin2003problems}, \citet{lafon2004diffusion},
\citet{bousquet2004mbr}, \citet{hein2005graphs},
\citet{hein2006uniform}, \citet{gine2006egl},
\citet{belkin2008towards}, \citet{vonluxburg2008csc}
have all shown various forms of asymptotic convergence for this graph
Laplacian. Although all of their results are encouraging, their results
do not apply to the random network models we study in this paper.
\citet{vu2010singular} studies how the singular vectors of a matrix
change under random perturbations.  These results are also encouraging.
However, the current paper uses a different method to study the
eigenvectors of the graph Laplacian.

\subsection{Statistical estimation}

Stochastic models are useful because they force us to think clearly
about the randomness in the data in a precise and possibly familiar
way. Many random network models have been proposed
[\citet{erdos1959random}, \citet{holland1981exponential},
\citet{holland1983stochastic}, \citet{frank1986markov},
\citet{watts1998collective}, \citet{barabasi1999emergence},
\citet{hoff2002latent}, \citet{van2004p2}, \citet
{goldenberg2009survey}.]
Some of these models, such as the Stochastic Blockmodel, have
well-defined communities. The Stochastic Blockmodel is characterized by the
fact that each node belongs to one of multiple blocks and the
probability of a relationship between two nodes depends only on the
block memberships of the two nodes. If the probability of an edge
between two nodes in the same block is larger than the probability of
an edge between two nodes in different blocks, then the blocks produce
communities in the random networks generated from the model.

Just as statisticians have studied when least-squares regression can
estimate the ``true'' regression model, it is natural and important for
us to study the ability of clustering algorithms to estimate the
``true'' clusters in a network model. Understanding when and why a
clustering algorithm correctly estimates the ``true'' communities would
provide a rigorous understanding of the behavior of these algorithms,
suggest which algorithm to choose in practice, and aid the
corroboration of algorithmic output.

This paper studies the performance of spectral clustering, a
nonparametric method, on a parametric task of estimating the blocks in
the Stochastic Blockmodel. It connects the first strain of clustering
research based on stochastic models to the second strain based on
heuristics and insights on network clusters. The stochastic blockmodel
allows for some first steps in understanding the behavior of
spectral clustering
and provides a benchmark to measure its performance. However, because
this model does not really account for the complexities observed in
several empirical
networks, good performance on the Stochastic Blockmodel should only be
considered
a necessary requirement for a good clustering algorithm.

Researchers have explored the performance of other clustering
algorithms under the Stochastic Blockmodel.
\citet{snijders1997estimation} showed the consistency under the
two block
Stochastic Blockmodel of a clustering routine that clusters the nodes
based on their degree distributions. Although this clustering is very
easy to compute it is not clear that the estimators would behave well
for larger graphs given the extensive literature on the long tail of
the degree distribution [\citet{albert2002statistical}]. Later,
\citet{condon1999algorithms} provided an algorithm and proved
that it is
consistent under the Stochastic Blockmodel, or what they call the
planted $\ell$-partition model. Their algorithm runs in linear time.
However, it always estimates clusters that contain an equal number of
nodes. More recently, \citet{bickel2009nonparametric} proved that under
the Stochastic Blockmodel, the maximizers of the Newman--Girvan
modularity [\citet{newman2004finding}] and what they call the
likelihood modularity are asymptotically consistent estimators of block
partitions. These modularities are objective functions that have no
clear relationship to the Girvan--Newman algorithm. Finding the maximum
of the modularities is NP hard [\citet{brandes2007modularity}]. It
is important to note that all aforementioned clustering results
involving the Stochastic Blockmodel are asymptotic in the number of
nodes, with a fixed number of blocks.\looseness=-1

The work of \citet{leskovec2008statistical} shows that in a
diverse set
of large empirical networks (tens of thousands to millions of nodes),
the size of the ``best'' clusters is not very large, around 100 nodes.
Modern applications of clustering require an asymptotic regime that
allows these sorts of clusters. Under the asymptotic regime cited in
the previous paragraph, the size of the clusters grows linearly with
the number of nodes. It would be more appropriate to allow the number
of communities to grow with the number of nodes. This restricts the
blocks from becoming too large, following the empirical observations of
\citet{leskovec2008statistical}.

This paper provides the first asymptotic clustering results that allow
the number of blocks in the Stochastic Blockmodel to grow with the
number of nodes. Similar to the asymptotic results on regression
techniques that allow the number of predictors to grow with the number
of nodes, allowing the number of blocks to grow makes the problem one
of high-dimensional learning. Following our initial technical report,
\citet{choi2010stochastic} also studied community detection under the
Stochastic Blockmodel with a~growing number of blocks. They used a
likelihood-based approach, which is computationally difficult to
implement. However, they are able to greatly weaken the assumptions of
this paper.

The Stochastic Blockmodel is an example of the more general latent
space model [\citet{hoff2002latent}] of a random network. Under
the latent space model, there are latent i.i.d. vectors $z_1,\ldots,
z_n$; one for each node. The probability that an edge appears between
any two nodes $i$ and~$j$ depends only on $z_i$ and $z_j$ and is
independent of all other edges and unobserved vectors. The results of
Aldous and Hoover show that this model characterizes the distribution
of all infinite random graphs with exchangeable nodes [\citet
{kallenberg2005probabilistic}]. The graphs with $n$ nodes generated
from a~latent space model can be viewed as a subgraph of an infinite
graph. In order to study spectral clustering under the Stochastic
Blockmodel, we first show that under the more general latent space
model, as the number of nodes grows, the eigenvectors of $L$, the
normalized graph Laplacian, converge to eigenvectors of the
``population'' normalized graph Laplacian that is constructed with
a~similarity matrix $\E(W| z_1,\ldots, z_n)$ (whose $i,j$th element is
the probability of an edge between node $i$ and $j$) taking the place
of the adjacency matrix $W$ in (\ref{Ldef1}). In many ways, $\E
(W| z_1,\ldots, z_n)$ is similar to the similarity matrix $K$ discussed
above, only this time the vectors ($z_1,\ldots, z_n$) and their
similarity matrix $\E(W| z_1,\ldots, z_n)$ are unobserved.

The convergence of the eigenvectors has implications beyond spectral
clustering. Graph visualization is an important tool for social network
analysts looking for structure in networks and the eigenvectors of the
graph Laplacian are an essential piece of one visualization technique
[\citet{koren2005drawing}]. Exploratory graph visualization allows
researchers to find structure in the network; this structure could be
communities or something more complicated
[\citet{liotta-graph}, \citet{freeman2000visualizing},
\citet{wasserman1994social}]. In terms of
the latent space model, if $z_1,\ldots, z_n$ form clusters or have some
other structure in the latent space, then we might recover this
structure from the observed graph using graph visualization. Although
there are several visualization techniques, there is very little
theoretical understanding of how these techniques perform under
stochastic models of structured networks. Because the eigenvectors of
the normalized graph Laplacian converge to ``population'' eigenvectors,
this provides support for a visualization technique similar to the one
proposed in \citet{koren2005drawing}.

The rest of the paper is organized as follows. The next subsection of
the \hyperref[sec1]{Introduction} give some preliminary definitions.
Following the
\hyperref[sec1]{Introduction}, there are four main sections; Section
\ref{sec2} studies the
latent space model, Section \ref{sec3} studies the Stochastic
Blockmodel as a
special case, Section \ref{sec4} presents some simulation results, and
Section \ref{sec5}
investigates the plausibility of a key assumption in five empirical
social networks. Section \ref{sec2} covers the eigenvectors of $L$
under the latent space model. The main technical result is Theorem~\ref
{frobeniusnormtheorem} in Section \ref{sec2}, which shows that, as the number
of nodes grows, the normalized graph Laplacian multiplied by itself
converges in Frobenius norm to a symmetric version of the population
graph Laplacian multiplied by itself. The Davis--Kahan theorem then
implies that the eigenvectors of these matrices are close in an
appropriate sense. Lemma \ref{square} specifies how the eigenvectors of
a matrix multiplied by itself are closely related to the eigenvectors
of the original matrix. Theorem \ref{eigenconvergence} combines Theorem
\ref{frobeniusnormtheorem} with the Davis--Kahan theorem and Lemma
\ref{square} to show that the eigenvectors of the normalized
graph
Laplacian converge to the population eigenvectors. Section~\ref{sec3} applies
these results to the high-dimensional Stochastic Blockmodel. Lemma \ref
{mu} shows that the population version of spectral clustering can
correctly identify the blocks in the Stochastic Blockmodel. Theorem
\ref{growTheorem} extends this result to the sample version of spectral
clustering. It uses Theorem \ref{eigenconvergence} to bound the number
of nodes that spectral clustering ``misclusters.'' This section
concludes with two examples. Section \ref{sec4} presents three
simulations that investigate how the asymptotic results apply to finite
samples. These simulations suggest an area for future research. The
main theorems in this paper require a strong assumption on the degree
distribution. Section \ref{sec5} investigates the plausibility of this
assumption with five empirical online social networks.
The discussion in Section \ref{discussion} concludes the paper.

\subsection{Preliminaries} \label{prelim}
The latent space model proposed by \citet{hoff2002latent} is a
class of
a probabilistic model for $W$.
\begin{definition}
For i.i.d. random vectors $z_1,\ldots, z_n \in\R^k$ and random
adjacency matrix $W \in\{0,1\}^{n \times n}$, let $\bP(W_{ij} |z_i,
z_j)$ be the probability mass function of $W_{ij}$ conditioned on $z_i$
and $z_j$.
If a probability distribution on $W$ has the conditional independence
relationships
\[
\bP(W| z_1,\ldots, z_n) = \prod_{i < j}\bP(W_{ij} |z_i, z_j)
\]
and $\bP(W_{ii} = 0) = 1$ for all $i$, then it is called an undirected
\textit{latent space model}.
\end{definition}

This model is often simplified to assume $\bP(W_{ij} |z_i, z_j) = \bP
(W_{ij} |\operatorname{dist}(z_i, z_j))$ where $\operatorname{dist}(\cdot, \cdot)$ is some
distance function. This allows the ``homophily by attributes''
interpretation that edges are more likely to appear between nodes whose
latent vectors are closer in the latent space.

Define $Z \in\R^{n\times k}$ such that its $i$th row is $z_i$ for all
$i\in V$. \textit{Throughout this paper, we assume $Z$ is fixed and
unknown}. Because $\bP(W_{ij} = 1|Z) = \E(W_{ij}|Z)$, the model is then
completely parametrized by the matrix
\[
\popw= \E(W|Z) \in\R^{n \times n},
\]
where $\popw$ depends on $Z$, but this is dropped for notational convenience.

The Stochastic Blockmodel, introduced by \citet{holland1983stochastic},
is a specific latent space model with well-defined communities. We use
the following definition of the undirected Stochastic Blockmodel:
\begin{definition}
The \textit{Stochastic Blockmodel} is a latent space model with
\[
\popw= Z B Z^T,
\]
where $Z \in\{0,1\}^{n \times k}$ has exactly one 1 in each row and at
least one 1 in each column and $B \in[0,1]^{k \times k}$ is full rank
and symmetric.
\end{definition}

We refer to $\popw$, the matrix which completely parametrizes the
latent space model, as the population version of $W$. Define population
versions of $L$ and~$D$ both in $\R^{n \times n}$ as
%
%
\begin{eqnarray}
\label{popdef}
\popd_{ii} &=& \sum_k \popw_{ik} ,\nonumber\\[-8pt]\\[-8pt]
\popl& = & \popd^{-1/2} \popw\popd^{-1/2},
\nonumber
\end{eqnarray}
where $\popd$ is a diagonal matrix, similar to before.

The results in this paper are asymptotic in the number of nodes $n$.
When it is appropriate, the matrices above are given a superscript of
$n$ to emphasize this dependence. Other times, this superscript is
discarded for notational convenience.

\section{Consistency under the latent space model}\label{sec2}

We will show that the empirical eigenvectors of $\Ln$ converge in the
appropriate sense to the population eigenvectors of $\popln$. If $\Ln$
converged to $\popln$ in Frobenius norm, then the Davis--Kahan theorem
would give the desired result. However, these matrices do not converge.
This is illustrated in an example below. Instead, we give a novel
result showing that under certain conditions $\Ln\Ln$ converges to
$\popln\popln$ in Frobenius norm. This implies that the eigenvectors
of $\Ln\Ln$ converge to the eigenvectors of $\popln\popln$. The
following lemma shows that these eigenvectors can be chosen to imply
the eigenvectors of~$\Ln$ converge to the eigenvectors of~$\popln$.
\begin{lemma}\label{square}
When $M \in\R^{n\times n}$ is a symmetric real matrix,
\begin{longlist}[(1)]
\item[(1)]$\lambda^2$ is an eigenvalue of $MM$ if and only if $\lambda$ or
$-\lambda$ is an eigenvalue of $M$.
\item[(2)] If $Mv = \lambda v$, then $MMv = \lambda^2 v$.
\item[(3)] Conversely, if $MM v = \lambda^2 v$, then $v$ can be written as a
linear combination of eigenvectors of $M$ whose eigenvalues are
$\lambda
$ or $-\lambda$.
\end{longlist}
\end{lemma}

A proof of Lemma \ref{square} can be found in Appendix \ref
{latentSpaceAppendix}.
\begin{Example*} To see how squaring a matrix helps convergence, let
the matrix $W \in\R^{n\times n}$ have i.i.d. $\operatorname{Bernoulli}(1/2)$
entries. Because the diagonal elements in $D$ grow like $n$, the matrix
$W/n$ behaves similarly to $D^{-1/2}WD^{-1/2}$.
Without squaring the matrix, the Frobenius distance from the matrix to
its expectation is
\[
\|W/n - \E(W)/n\|_F = \frac{1}{n}\sqrt{\sum_{i, j} \bigl(W_{ij} - \E
(W_{ij})\bigr)^2} = 1/2.
\]
Notice that, for $i\ne j$,
\[
[WW]_{ij} = \sum_k W_{ik}W_{kj} \sim\operatorname{Binomial}(n, 1/4)
\]
and $[WW]_{ii} \sim\operatorname{Binomial}(n,1/2)$. So, for any $i,j$,
$[WW]_{ij} - \E[WW]_{ij} =\break o(n^{1/2}\times\log n)$.
Thus, the Frobenius distance from the squared matrix to its expectation is
\[
\|WW/n^2 - \E(WW)/n^2\|_F = \frac{1}{n^2} \sqrt{\sum_{i, j}
([WW]_{ij} - \E[WW]_{ij})^2} = o\biggl(\frac{\log n}{
n^{1/2}}\biggr).
\]
\end{Example*}

When the elements of $W$ are i.i.d. $\operatorname{Bernoulli}(1/2)$, $(W/n)^2$
converges in Frobenius norm and $W/n$ does not. The next theorem
addresses the convergence of $\Ln\Ln$.

Define
%
%
\begin{equation} \label{tau}
\tau_n = \min_{i = 1,\ldots, n} \popdni/ n.
\end{equation}
Recall that $\popdni$ is the expected degree for node $i$. So, $\tau_n$
is the minimum expected degree, divided by the maximum possible degree.
It measures how quickly the number of edges accumulates.
\begin{theorem} \label{frobeniusnormtheorem}
Define the sequence of random matrices $\Wn\in\{0,1\}^{n \times n}$
to be from a sequence of latent space models with population matrices
$\popwn\in[0,1]^{n \times n}$. With $\Wn$, define the observed graph
Laplacian $\Ln$ as in (\ref{Ldef1}). Let $\popln$ be the population
version of $\Ln$ as defined in (\ref{popdef}). Define $\tau_n$
as in (\ref{tau}).

If there exists $N>0$, such that $\tau_n^2 \log n > 2$ for all $n >
N$, then
\[
\bigl\| \Ln\Ln- \popln\popln\bigr\|_F = o\biggl( \frac{\log n}{\tau_n^{2} n^{
1/2}} \biggr) \qquad\mbox{a.s.}
\]
\end{theorem}

Appendix \ref{latentSpaceAppendix} contains a nonasymptotic bound on
$\| \Ln\Ln- \popln\popln\|_F$ as well as the proof of Theorem \ref
{frobeniusnormtheorem}. The main condition in this theorem is the lower
bound on $\tau_n$. This sufficient condition is used to produce
Gaussian tail bounds for each of the $D_{ii}$ and other similar quantities.

For any symmetric matrix $M$, define $\lambda(M)$ to be the eigenvalues
of $M$ and for any interval $S \subset\R$, define
\[
\lambda_S(M) =\{ \lambda(M) \cap S\}.
\]
Further, define $\bar\lambda_1^{(n)}\ge\cdots\ge\bar\lambda
_n^{(n)}$ to be the elements of $\lambda(\popln\popln)$ and $
\lambda_1^{(n)}\ge\cdots\ge\lambda_n^{(n)}$ to be the elements of $\lambda
(\Ln\Ln)$.
The eigenvalues of $\Ln\Ln$ converge in the following sense,
%
%
\begin{eqnarray} \label{evbound}
\max_i \bigl| \lambda^{(n)}_i - \bar\lambda^{(n)}_i\bigr| &\le& \bigl\|\Ln\Ln-
\popln\popln\bigr\|_F \nonumber\\[-8pt]\\[-8pt]
& = & o\biggl( \frac{\log n}{\tau_n^{2} n^{ 1/2}} \biggr) \qquad\mbox{a.s.}\nonumber
\end{eqnarray}
This follows from Theorem \ref{frobeniusnormtheorem}, Weyl's inequality
[\citet{bhatia1987perturbation}], and the fact that the Frobenius
norm is an upper bound of the spectral norm.

This shows that under certain conditions on $\tau_n$, the eigenvalues
of $\Ln\Ln$ converge to the eigenvalues of $ \popln\popln$. In order
to study spectral clustering, it is now necessary to show that the
eigenvectors also converge. The Davis--Kahan theorem provides a bound
for this.
\begin{prop}[(Davis--Kahan)]\label{DavisKahan}  Let $S \subset
\R
$ be an interval. Denote $\mathcal{X}$ as an orthonormal matrix whose
column space is equal to the eigenspace of $\popl\popl$ corresponding
to the eigenvalues in $\lambda_S(\popl\popl)$ [more formally, the
column space of $\mathcal{X}$ is the image of the spectral projection
of $\popl\popl$ induced by $\lambda_S(\popl\popl)$]. Denote by $X$
the analogous quantity for $LL$. Define the distance between $S$ and
the spectrum of $\popl\popl$ outside of $S$ as
\[
\delta= \min\{ |\ell- s|; \ell\mbox{ eigenvalue of } \popl\popl,
\ell\notin S, s \in S\}.
\]
If $\mathcal{X}$ and $X$ are of the same dimension, then there is an
orthonormal matrix~$O$, that depends on $\mathcal{X}$ and $X$, such that
\[
\frac{1}{2}\| X - \mathcal{X} O \|_F^2 \le\frac{\|L L - \popl\popl
\|
_F^2}{\delta^2}.
\]
\end{prop}

The original Davis--Kahan theorem bounds the ``canonical angle,'' also
known as the ``principal angle,'' between the column spaces of
$\mathcal
{X}$ and $X$. Appendix \ref{dkappendix} explains how this can be
converted into the bound stated above. To understand why the
orthonormal matrix $O$ is included, imagine the situation that $L =
\popl$. In this case $X$ is not necessarily equal to $\popx$. At
a~minimum, the columns of $X$ could be a permuted version of those in
$\popx$. If there are any eigenvalues with multiplicity greater than
one, these problems could be slightly more involved. The matrix $O$
removes these inconveniences and related inconveniences.

The bound in the Davis--Kahan theorem is sensitive to the value $\delta
$. This reflects that when there are eigenvalues of $\popl\popl$ close
to $S$, but not inside of $S$, then a small perturbation can move these
eigenvalues inside of $S$ and drastically alter the eigenvectors. The
next theorem combines the previous results to show that the
eigenvectors of $\Ln$ converge to the eigenvectors of $\popln$. Because
it is asymptotic in the number of nodes, it is important to allow $S$
and $\delta$ to depend on $n$. For a sequence of open intervals\vadjust{\eject} $S_n\subset\R$, define
%
%
\begin{eqnarray}
\label{delta1} \delta_n &=& \inf\bigl\{ |\ell- s|; \ell\in\lambda\bigl(
\popln\popln\bigr) , \ell\notin S_n, s \in S_n\bigr\}, \\
\label{delta2} \delta_n' &=& \inf\bigl\{ |\ell- s|; \ell\in\lambda
_{S_n}\bigl(\popln\popln\bigr) , s \notin S_n\bigr\}, \\
\label{Sprimedef}
S_n' &=& \{ \ell; \ell^2 \in S_n\}.
\end{eqnarray}
The quantity $\delta_n'$ is added to measure how well $S_n$ insulates
the eigenvalues of interest. If $\delta_n'$ is too small, then some
important empirical eigenvalues might fall outside of $S_n$. By
restricting the rate at which $\delta_n$ and $\delta_n'$ converge to
zero, the next theorem ensures the dimensions of $X$ and $\popx$ agree
for a~large enough $n$. This is required in order to use the
Davis--Kahan theorem.
\begin{theorem} \label{eigenconvergence}
Define $\Wn\in\{0,1\}^{n \times n}$ to be a sequence of growing
random adjacency matrices from the latent space model with population
matrices~$\popwn$. With $\Wn$, define the observed graph Laplacian
$\Ln
$ as in (\ref{Ldef1}). Let~$\popln$ be the population version of~$\Ln$
as defined in (\ref{popdef}).
Define $\tau_n$ as in~(\ref{tau}). With a sequence of open
intervals $S_n \subset\R$, define $\delta_n$, $\delta_n'$ and $S_n'$
as in~(\ref{delta1}), (\ref{delta2}) and (\ref{Sprimedef}).

Let $k_n = |\lambda_{S_n'}( \Ln)|$, the size of the set $\lambda
_{S_n'}( \Ln)$. Define the matrix $X_n \in\R^{n \times k_n}$ such that
its orthonormal columns are the eigenvectors of symmetric matrix $\Ln$
corresponding to all the eigenvalues contained in $\lambda_{S_n'}( \Ln)$.
For $\mathscr{K}_n = |\lambda_{S_n'}( \popln)|$, define $\popx_n
\in\R
^{n \times\mathscr{K}_n}$ to be the analogous matrix for symmetric
matrix $\popln$ with eigenvalues in $\lambda_{S_n'}( \popln)$.

Assume that $n^{-1/2} (\log n)^2 = O(\min\{\delta_n, \delta_n'\})$.
Also assume that there exists positive integer $N$ such that for all $n
> N$, it follows that $\tau_n^2 > 2/ \log n$.

Eventually, $k_n = \mathscr{K}_n$. Afterward, for some sequence of
orthonormal rotations~$O_n$,
\[
\| X_n - \popx_n O_n \|_F = o\biggl( \frac{\log n}{\delta_n \tau_n^{2}
n^{ 1/2}} \biggr) \qquad\mbox{a.s.}
\]
\end{theorem}

A proof of Theorem \ref{eigenconvergence} is in Appendix \ref
{proofeigenconvergence}. There are two key assumptions in Theorem \ref
{eigenconvergence}:
\begin{eqnarray*}
\mbox{(1) }\quad n^{-1/2} (\log n)^2 &=& O(\min\{\delta_n, \delta_n'\}), \\
\mbox{(2) }\hspace*{60.5pt}\tau_n^2 &>& 2/ \log n.
\end{eqnarray*}
The first assumption ensures that the ``eigengap,'' the gap between the
eigenvalues of interest and the rest of the eigenvalues, does not
converge to zero too quickly. The theorem is most interesting when $S$
includes only the leading eigenvalues. This is because the eigenvectors
with the largest eigenvalues have the potential to reveal clusters or
other structures in the network. When these leading eigenvalues are
well separated from the smaller eigenvalues, the eigengap is large. The
second assumption ensures that the expected degree of each node grows
sufficiently fast. If $\tau_n$ is constant, then the expected degree of
each node grows linearly. The assumption $\tau_n^2 > 2/ \log n$ is
almost as restrictive.

The usefulness of Theorem \ref{eigenconvergence} depends on how well
the eigenvectors of~$\popln$ represent the characteristics of interest
in the network. For example, under the Stochastic Blockmodel with $B$
full rank, if~$S_n$ is chosen so that~$S_n'$ contains all nonzero
eigenvalues of $\popln$, then the block structure can be determined
from the columns of $\mathcal{X}_n$. It can be shown that nodes $i$
and~$j$ are in the same block if and only if the $i$th row of $\mathcal
{X}_n$ equals the $j$th row. The next section examines how spectral
clustering exploits this structure, using~$X_n$ to estimate the block
structure in the Stochastic Blockmodel.

\section{The Stochastic Blockmodel} \label{sec3}
The work of \citet{leskovec2008statistical} shows that the sizes
of the
best clusters are not very large in a diverse set of empirical
networks, suggesting that the appropriate asymptotic framework should
allow for the number of communities to grow with the number of nodes.
This section shows that, under suitable conditions, spectral clustering
can correctly partition most of the nodes in the Stochastic Blockmodel,
even when the number of blocks grows with the number of nodes.

The Stochastic Blockmodel, introduced by \citet{holland1983stochastic},
is a specific latent space model. Because it has well-defined
communities in the model, community detection can be framed as a
problem of statistical estimation. The important assumption of this
model is that of stochastic equivalence within the blocks; if two nodes
$i$ and $j$ are in the same block, rows $i$ and $j$ of $\popw$ are equal.

Recall in the definition of the undirected Stochastic Blockmodel,
\[
\popw= Z B Z^T,
\]
where $Z \in\{0,1\}^{n \times k}$ is fixed and has exactly one 1 in
each row and at least one~1 in each column and $B \in[0,1]^{k \times
k}$ is full rank and symmetric. In this definition there are $k$ blocks
and $n$ nodes. If the $i,g$th element of $Z$ equals one ($Z_{ig} = 1$),
then node $i$ is in block $g$. As before, $z_i$ for $i = 1,\ldots, n$
denotes the $i$th row of $Z$. The matrix $B \in[0,1]^{k \times k}$
contains the probability of edges within and between blocks. Some
researchers have allowed for $Z$ to be random, we have decided to focus
instead on the randomness of $W$ conditioned on $Z$. The aim of a
clustering algorithm is to estimate $Z$ (up to a permutation of the
columns) from $W$.

This section bounds the number of ``misclustered'' nodes. Because a
permutation of the columns of $Z$ is unidentifiable in the Stochastic
Blockmodel, it is not obvious what a ``misclustered'' node is. Before
giving our definition of ``misclustered,'' some preliminaries are needed
to explain why it is a reasonable definition.
The next paragraphs examine the behavior of spectral clustering applied
to the population graph Laplacian $\popl$. Then, this is compared to
spectral clustering applied to the observed graph Laplacian $L$. This
motivates our definition of ``misclustered.''

Recall that the spectral clustering algorithm applied to $L$,
\begin{enumerate}[(3)]
\item[(1)] finds the eigenvectors, $X \in R^{n \times k}$,
\item[(2)] treats each row of the matrix $X$ as a point in $\R^k$, and
\item[(3)] runs $k$-means on these points.
\end{enumerate}
$k$-means is an objective function. Applied to the points $\{x_1,\ldots
, x_n\} \subset R^k$ it is \citet{steinhaus1956division},
%
%
\begin{equation} \label{kmeansdef}
\min_{\{ m_1,\ldots, m_k\} \subset\R^k} \sum_i \min_g \|x_i -
m_g\|_2^2.
\end{equation}
The analysis in this paper addresses the true optimum of (\ref
{kmeansdef}). (In practice, this optimization problem can suffer from
local optima.) The vectors $m_1^*,\ldots, m_k^*$ that optimize the
$k$-means function are referred to as the \textit{centroids} of the $k$
clusters.

This next lemma shows that spectral clustering applied to the
population Laplacian, $\popl$, can discover the block structure in the
matrix $Z$. This lemma is essential to defining ``misclustered.''
\begin{lemma} \label{mu}
Under the Stochastic Blockmodel with $k$ blocks,
\[
\popw= Z B Z^T \in R^{n \times n} \qquad\mbox{for $B \in R^{k \times k}$
and $Z\in\{0,1\}^{n \times k}$,}
\]
define $\popl$ as in (\ref{popdef}). There exists a matrix $\mu\in
R^{k \times k}$ such that the columns of $Z \mu$ are the eigenvectors
of $\popl$ corresponding to the nonzero eigenvalues.
Further,
%
%
\begin{equation}\label{zeq}
z_i \mu= z_j \mu\quad\Leftrightarrow\quad z_i = z_j ,
\end{equation}
where $z_i$ is the $i$th row of $Z$.
\end{lemma}

A proof of Lemma \ref{mu} is in Appendix \ref{sbmappendix}.

Equivalence statement (\ref{zeq}) implies that under the $k$ block
Stochastic Blockmodel there are $k$ unique rows in the eigenvectors
$Z\mu$ of $\popl$. This has important consequences for the spectral
clustering algorithm.
The spectral clustering algorithm applied to $\popl$ will run $k$-means
on the rows of $Z \mu$. Because there are only $k$ unique points, each
of these points will be a centroid of one of the resulting clusters.
Further, if $z_i \mu= z_j \mu$, then $i$ and $j$ will be assigned to
the same cluster. With equivalence statement (\ref{zeq}), this implies
that spectral clustering applied to the matrix $\popl$ can perfectly
identify the block memberships in $Z$. Obviously, $\popl$ is not
observed. In practice, spectral clustering is applied to $L$. Let $X
\in R^{n \times k}$ be a matrix whose orthonormal columns are the
eigenvectors corresponding to the largest $k$ eigenvalues (in absolute
value) of $L$.
\begin{definition} \label{cdef}
Spectral clustering\vspace*{1pt} applies the $k$-means algorithm to the rows of
$X$, that is, each row is a point in $R^k$. Each row is assigned to one
cluster and each of these clusters has a centroid. Define $c_1,\ldots,
c_n \in R^k$ such that $c_i$ is the the centroid corresponding to the
$i$th row of $X$.
\end{definition}

Recall that $z_i \mu$ is the centroid corresponding to node $i$ from
the population analysis. If the observed centroid $c_i$ is closer to
the population centroid~$z_i \mu$ than it is to any other population
centroid $z_j \mu$ for $z_j \ne z_i$, then it appears that node $i$ is
correctly clustered. This definition is appealing because it removes
some of the cluster identifiability problem. However, the eigenvectors
add one additional source of undentifiability.
Let $O \in R^{k \times k}$ be the orthonormal rotation from Theorem
\ref
{eigenconvergence}.
Consider node $i$ to be correctly clustered if, $c_i$ is closer to $z_i
\mu O$ than it is to any other (rotated) population centroid~$z_j \mu
O$ for $z_j \ne z_i$. The slight complication with $O$ stems from the
fact that the vectors $c_1,\ldots, c_n$ are constructed from the
eigenvectors in $X$ and Theorem~\ref{eigenconvergence} shows these
eigenvectors converge to the \textit{rotated} population eigenvectors:
$\popx O = Z \mu O$.

Define $P$ to be the population of the largest block in $Z$.
%
%
\begin{equation}\label{pdef}
P = \max_{j = 1,\ldots, k} (Z^T Z)_{jj}.
\end{equation}
The following provides a sufficient condition for a node to be
correctly clustered.
\begin{lemma} \label{misclusteredLemma}
For the orthonormal matrix $O \in\R^{k \times k}$ from Theorem \ref
{eigenconvergence},
%
%
\begin{eqnarray}\label{suff}
&&\|c_i - z_i \mu O\|_2 < 1/\sqrt{2 P}
\\
%
%
\label{rightclust}
&&\hspace*{4pt}\quad\Longrightarrow\quad\|c_i - z_i \mu O \|_2 < \|c_i - z_j \mu
O\|_2\qquad
\mbox{for any } z_j \ne z_i.
\end{eqnarray}
\end{lemma}

A proof of Lemma \ref{misclusteredLemma} is in Appendix \ref{sbmappendix}.

Line (\ref{rightclust}) is the previously motivated definition of
correctly clustered. Thus, Lemma \ref{misclusteredLemma} shows that the
inequality in line (\ref{suff}) is a sufficient condition for node $i$
to be correctly clustered.
\begin{definition}\label{misclustereddef}
Define the set of misclustered nodes as the nodes that do not satisfy
the sufficient condition (\ref{suff}),
%
%
\begin{equation} \label{mdef}
\mathscr{M} = \bigl\{ i \dvtx\|c_i - z_i \mu O\|_2 \ge1/\sqrt{2P}
\bigr\}.
\end{equation}
\end{definition}

The next theorem bounds the size of the set $\mathscr{M}$.
\begin{theorem}\label{growTheorem}
Suppose $W \in\R^{n \times n}$ is an adjacency matrix from the
Stochastic Blockmodel with $k_n$ blocks. Define the population graph
Laplacian,~$\popl$, as in (\ref{popdef}). Define $|\bar\lambda_1|
\ge
| \bar\lambda_2| \ge\cdots\ge|\bar\lambda_{k_n}| >0$ as the
absolute values of the $k_n$ nonzero eigenvalues of $\popl$.
Define $\mathscr{M}$, the set of misclustered nodes, as in (\ref
{mdef}). Define $\tau_n$ as in (\ref{tau}) and assume there exists $N$
such that for all $n >N$, $\tau_n^2 > 2/\log n$. Define $P_n$ as in
(\ref{pdef}).
If $n^{-1/2} (\log n)^2 = O(\lambda_{k_n}^2)$, then the number of
misclustered nodes is bounded
\[
|\mathscr{M}| = o\biggl(\frac{P_n (\log n)^2}{ \lambda_{k_n}^4 \tau_n^4
n }\biggr).
\]
\end{theorem}

A proof of Theorem \ref{growTheorem} is in Appendix \ref{sbmappendix}.
The two main assumptions of Theorem \ref{growTheorem} are
\begin{eqnarray*}
\mbox{(1)}\hspace*{41.5pt}\quad n ^{-1/2} (\log n)^2 &=& O(\lambda_{k_n}^2),\\
\mbox{(2)}\quad\mbox{eventually}\qquad \tau_n^2 \log n &>&2 .
\end{eqnarray*}
They imply the conditions needed to apply Theorem \ref
{eigenconvergence}. The first assumption requires that the smallest
nonzero eigenvalue of $\popl$ is not too small. Combined with an
appropriate choice of $S_n$, this assumption implies the eigengap
assumption in Theorem \ref{eigenconvergence}. The second assumption is
exactly the same as the second assumption in Theorem \ref
{eigenconvergence}. Section \ref{sec4} investigates the sensitivity of spectral
clustering to these two assumptions. Section \ref{sec5} examines the
plausibility of assumption (2) on five empirical online social networks.

In all previous spectral clustering algorithms, it has been suggested
that the eigenvectors corresponding to the largest eigenvalues reveal
the clusters of interest. The above theorem suggests that before
finding the largest eigenvalues, you should first order them by
absolute value. This allows for large and negative eigenvalues. In
fact, eigenvectors of $L$ corresponding to eigenvalues close to
negative one (all eigenvalues of $L$ are in $[-1,1]$) discover
``heterophilic'' structure in the network that can be useful for
clustering. For example, in the network of dating relationships in a
high school, two people of opposite sex are more likely to date than
people of the same sex. This pattern creates the two male and female
``clusters'' that have many fewer edges within than between clusters. In
this case, $L$ would likely have an eigenvalue close to negative one.
The corresponding eigenvector would reveal these ``heterophilic'' clusters.
\begin{Example*} To examine the ability of spectral clustering to
discover heterophilic clusters, imagine a Stochastic Blockmodel with
two blocks and two nodes in each block. Define
\[
B = \pmatrix{0 & 1 \cr1 & 0}.
\]
In this case, there are no connections within blocks and every member
is connected to the two members of the opposite block. There is no
variability in the matrix $W$. The rows and columns of $L$ can be
reordered so that it is a~block matrix. The two block matrices down the
diagonal are $2\times2$ matrices of zeros and all the elements in the
off diagonal blocks are equal to $1/2$. There are two nonzero
eigenvalues of~$L$. Any constant vector is an eigenvector of~$L$ with
eigenvalue equal to one. The remaining eigenvalue belongs to any
eigenvector that is a constant multiple of $(1,1,-1,-1)$. In this case,
with perfect ``heterophilic'' structure, the eigenvector that is useful
for finding the clusters has eigenvalue negative one.

Heuristically, the reason spectral clustering can discover these
heterophilic blocks is related to our method of proof. The $i,j$th
element of $WW$ is the number neighbors that nodes $i$ and $j$ have in
common. In both heterophilic and homophilic cases, if nodes $i$ and $j$
are in the same block, then they should have several neighbors in
common. Thus, $[WW]_{ij}$ is large. Similarly, $[LL]_{ij}$ is large.
This shows that the number of common neighbors is a measure of
similarity that is robust to the choice of hetero- or homophilic
clusters. Because spectral clustering uses a related measure of
similarity, it is able to detect both types of clusters.

In order to clarify the bound on $|\mathscr{M}|$ in Theorem \ref
{growTheorem}, a simple example illustrates how $\lambda_{k_n}$, $\tau
_n$ and $P$ might depend on $n$.
\end{Example*}
\begin{definition} \label{fourparameter} The \textit{four parameter
Stochastic Blockmodel} is parametri\-zed by $k, s, r$ and $p$. There
are $k$ blocks each containing $s$ nodes. The probability of a
connection between two nodes in two separate blocks is $r \in[0,1]$
and the probability of a connection between two nodes in the same block
is $p+r \in[0,1]$.
\end{definition}
\begin{Example*} In the four parameter Stochastic Blockmodel, there
are $n = ks$ nodes. Notice that $P_n = s$ and $\tau_n > r$. Appendix
\ref{sbmappendix} shows that the smallest nonzero eigenvalue of the
population graph Laplacian is equal to
\[
\lambda_k = \frac{1}{k(r/p) + 1}.
\]
Using Theorem \ref{growTheorem}, if $p \ne0$ and $k = O(n^{1/4}/\log
n)$, then
%
%
\begin{equation}\label{Mbound}
|\mathscr{M}| = o(k^3 (\log n)^2)\qquad\mbox{a.s.}
\end{equation}
Further, the proportion of nodes that are misclustered converges to zero,
\[
\frac{| \mathscr{M}|}{n} = o(n^{-1/4} ) \qquad\mbox{a.s.}
\]

This example is particularly surprising after noticing that if $k =
n^{\alpha}$ for $\alpha\in(0, 1/4)$, then the vast majority of edges
connect nodes in different blocks. To see this, look at a sequence of
models such that $k = n^{\alpha}$. Note that $s = n^{1 - \alpha}$. So,
for each node, the expected number of connections to nodes in the same
block is $(p+r)n^{1- \alpha}$ and the expected number of connections to
nodes in different blocks is $r(n - n^{1-\alpha})$.
\[
\frac{\mbox{Expected number of in block connections}}{\mbox{Expected
number of out of block connections}} = \frac{(p+r)n^{1-\alpha} }{r(n -
n^{1-\alpha})} = O(n^{-\alpha}).
\]
These are not the tight communities that many imagine when considering
networks. Instead, a dwindling fraction of each node's edges actually
connect to nodes in the same block. The vast majority of edges connect
nodes in different blocks.

A more refined result would allow $r$ to decay with $n$. However, when
$r$ decays, so does the minimum expected degree and the tail bounds
used in proving Theorem~\ref{frobeniusnormtheorem} requires the minimum
expected degree to grow nearly as fast as $n$. Allowing $r$ to decay
with $n$ is an area for future research.
\end{Example*}

\section{Simulations} \label{sec4}

$\!\!\!$Three simulations in this section illustrate how the asymptotic bounds
in this paper can be a guide for finite sample results. These
simulations emphasize the importance of the eigengap in Theorem \ref
{eigenconvergence} and suggest that the asymptotic bounds in this paper
hold for relatively small networks. The simulations also suggest two
shortcomings of the theoretical results in this paper. First,
Simulation 1 shows that spectral clustering appears to be consistent in
some situations. Unfortunately, the theoretical results in Theorem \ref
{growTheorem} are not sharp enough to prove consistency. Second,
Simulation 3 suggests that spectral clustering is still consistent even
when the minimum expected node degree grows more slowly than the number
of nodes. However, the theorems above require a stronger condition,
that the minimum expected degree grows almost linearly with the number
of nodes.

All data are simulated from the four parameter Stochastic Blockmodel
(Definition \ref{fourparameter}). In the first simulation, the number
of nodes in each block $s$ grows while the number of blocks $k$ and the
probabilities $p$ and $r$ remain fixed. In the second simulation, $k$
grows while $s,p$ and $r$ remain fixed. In the final simulation, $s$
and $k$ remain fixed while $r$ and $p$ shrink such that $p/r$ remains
fixed. Because $kr/p$ is fixed, the eigengap is also fixed.

There is one important detail to recreate our simulation results below.
The spectral clustering result stated in Theorem \ref{growTheorem},
requires the true optimum of the $k$-means objective function. This is
very difficult to ensure. However, only one step in the proof of
Theorem \ref{growTheorem} requires the true optimum. The optimum of
$k$-means satisfies inequality \ref{kmeansbound} in the Appendix~\ref
{sbmappendix}. In
simulations, this inequality can be verified directly. For the
simulations below, the $k$-means algorithm is run several times, all
with random initializations, until the bound \ref{kmeansbound} is
met.\vspace*{8pt}

\begin{figure}

\includegraphics{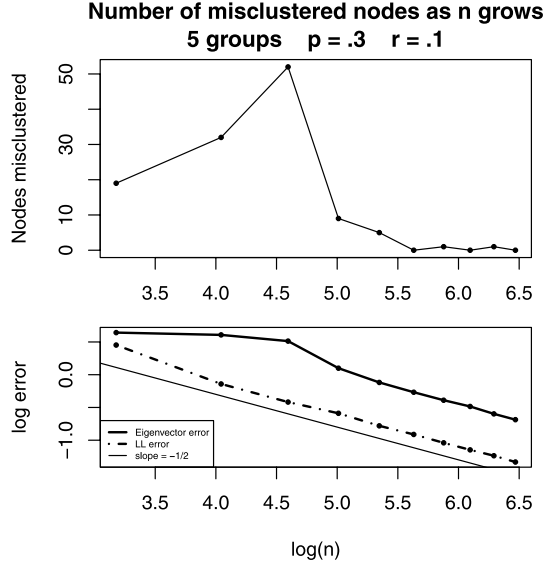}

\caption{The top panel in this figure displays the number of
misclustered nodes
plotted against $\log n$. The bottom panel displays both $\log\|LL -
\popl\popl\|_F$ and $\log\|X - \popx O\|_F$ plotted against $\log n$.
Each dot represents one simulation of the model. In addition, the
bottom panel has a line with slope $-1/2$. This figure illustrates two
things. First, after a certain threshold (around $\log n = 4.7$), the
eigenvectors of the graph Laplacian begin to converge and after this
point, the number of misclustered nodes converges to zero. Second, the
lines representing $\log\|LL - \popl\popl\|_F$ and $\log\|X - \popx
O\|_F$ are approximately parallel to the line with slope $-1/2$. This
suggests\vspace*{1pt} that they converge around rate $O(n^{-1/2})$, similar to the
theoretical results in Lemma \protect\ref{frobeniusnormtheorem} and Theorem
\protect\ref{eigenconvergence}.}
\label{ngrows}
\end{figure}

\textit{Simulation} 1: In this simulation, $k = 5, p = 0.2, r =
0.1$ and the number of members in each group grows from 8 to 215. This
implies that $n$ grows from 40 to 1075. Equation (\ref{Mbound})
suggests that the number of misclustered nodes should grow more slowly
than $(\log n)^2$. In fact, Figure \ref{ngrows} shows that once there
are enough nodes, the number of misclustered nodes converges to zero.
The top plot displays the number of misclustered nodes plotted against
$\log n$, which initially increases. Then, it falls precipitously.

%

%
%
%
\begin{figure}[b]

\includegraphics{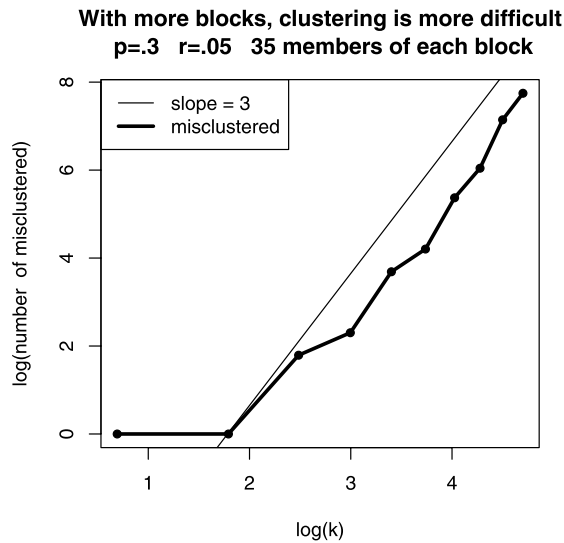}

\caption{This figure plots the number of miclustered nodes (thicker line)
against $\log k$. Each dot represents one simulation from the model.
Additionally, there is a line with slope~$3$ (thinner line). Equation
(\protect\ref{Mbound}) says that the number of misclustered nodes is
$o(k^3 \log k)$. Because the thicker line has a slope that is similar to the
thinner line, this result appears to be a good approximation.}
\label{kgrows}
\end{figure}

The lower plot in Figure \ref{ngrows} displays why the number of
misclustered nodes falls so precipitously. It plots $ \log\| LL -
\popl\popl\|_F$ (dashed bold line) and $\log\|X - \popx O\|_F$
(solid bold line) on the vertical axis against $\log n$ on the
horizontal axis. Also displayed in this plot is a line with slope $-1/2$
(solid thin line). Note that the solid bold line starts to run parallel
to the solid thin line once $\log n > 4.5$. After this point, the
eigenvectors converge, and spectral clustering begins to correctly
cluster all of the nodes.
The proof of the convergence of the eigenvectors for Theorem \ref
{eigenconvergence}, requires an eigengap condition,\looseness=-1
\[
n^{-1/2} \log n = O(\min\{\delta_n, \delta_n'\}).
\]\looseness=0
Similar to the example in the previous section, $S_n$ can be chosen
in this four parameter model so that $\min\{\delta_n, \delta_n'\} =
(k(r/p) + 1)^{-2}$.\vadjust{\eject} In this simulation, the eigenvectors begin to
converge, and the number of miclustered nodes drops just after the
bound $n^{-1/2} < (k(r/p) + 1)^{-2}$ is met. Ignoring the $\log n $
factor, this suggests that the eigengap condition in Theorem \ref
{eigenconvergence} is necessary.

This simulation demonstrates the importance of the relationship between
the sample size and the eigengap. In this simulation, there needs to be
roughly 50~nodes in each block to separate the informative eigenvectors
from the uninformative eigenvectors. Once there are enough nodes, the
empirical eigenvectors are close to the population eigenvectors. Then,
spectral clustering can estimate the block structure.

The lower plot in Figure \ref{ngrows} also suggests that, ignoring
$\log n$ factors, the rates of convergence given in Theorem \ref
{frobeniusnormtheorem} and Theorem \ref{eigenconvergence} are sharp.
Both~$LL$ and the eigenvectors $X$ converge at a rate $O(n^{-1/2})$.
This is because the the dashed bold line and the solid bold line (for
large enough $n$) are approximately parallel to the solid thin
line.\vspace*{8pt}

\textit{Simulation} 2: In this simulation from the four parameter
Stochastic Blockmodel, each block contains 35 nodes, $p = 0.3$ and $r =
0.05$. The number of blocks $k$ grows from 2 to 110. Equation (\ref
{Mbound}) suggests that under this asymptotic regime, the number of
misclustered nodes should grow more slowly than $k^3 (\log n)^2$.
Figure~\ref{kgrows} shows how this theoretical quantity can be an
appropriate guide.

Figure \ref{kgrows} plots the $\log$ of the number of misclustered
nodes (bold line) against $\log k$. For comparison, a line with slope 3
is also plotted (thin line). Because the bold line has a slope
approximately equal to the thin line, the number of misclustered nodes
is approximate to $k^3$.

This simulation demonstrates that as the number of blocks grows, the
number of misclustered nodes also grows. Although $\| LL - \popl\popl
\|_F$ converges under this asymptotic regime, $\|X - \popx O\|_F$ does
not because the eigengap shrinks more quickly than the number of nodes
can tolerate.\vspace*{8pt}

\textit{Simulation} 3: The theorems in this paper assume that the
smallest expected degree grows close to linearly with the number of
nodes in the graph. This simulation examines the sensitivity of
spectral clustering to this assumption. Recall that the smallest
expected degree is equal to $n \tau$.

In this simulation, there are three different designs all from the four
parameter Stochastic Blockmodel. Each design has three blocks ($k =
3$). One design contains 50 nodes in each block, another contains 150
in each block, and the last design contains 250 nodes in each block. To
investigate how sensitive spectral clustering is to the value of $\tau
= p/k + r$, the probabilities~$p$ and $r$ must change.\vspace*{2pt} However, to
isolate the effect of $\tau$ from the effect of the eigengap $(k (r/p)
+ 1)^{-2}$, it is necessary to keep the ratio $p/r$ constant. Fixing
$p/r = 2$ ensures that the eigengap is fixed at $4/25$.

\begin{figure}

\includegraphics{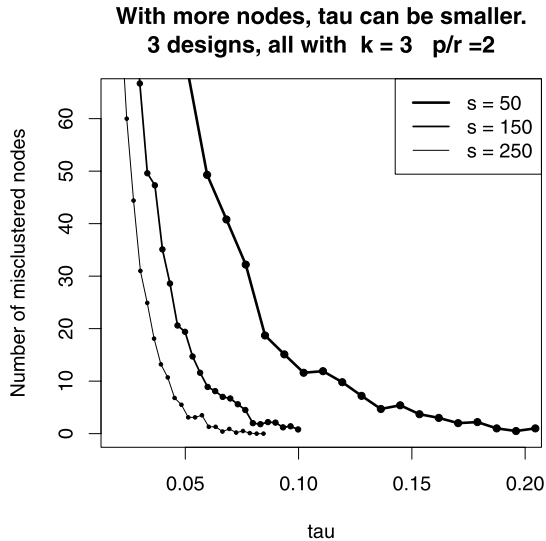}

\caption{This figure displays the number of misclustered nodes from three
different models plotted against $\tau= \min_i E(D_{ii})/n$. The first
model has $50$ nodes in each block (thickest line), the second model
has $150$ nodes in each block (line with medium thickness), the third
model has $250$ nodes in each block (thinnest line). Each dot
represents the average of ten simulations from the model. In each of
these models, $p$ and $r$ decrease such that $p/r$ is always equal to
$2$. This ensures that $\tau$ goes to zero, while the eigengap remains
constant. Each of the three models is sensitive to small values of~$\tau
$. However, the larger models can tolerate a smaller value of $\tau$.
This suggests that as $n$ increases, $\tau$~should be allowed to
decrease. The theorems in this paper do not allow for that possibility.}
\label{tauShrink}
\end{figure}

The results for Simulation 3 are displayed in Figure \ref{tauShrink}.
%
%
The value $\tau$ is on the horizontal axis, and the number of
misclustered nodes is on the vertical axis. There are three lines. The
thickest line represents the design with 50 nodes in each block. The
line of medium thickness represents the design with 150 nodes in each
block. The thinnest line represents the design with 250 nodes in each
block. All three lines increase as $\tau$ approaches zero (reading the
figure from right to left). The thickest line starts to increase at
$\tau= 0.20$. The thinnest line starts to increase at $\tau= 0.07$. The
line with medium thickness increases in between these two lines.

Because the thinner lines start to increase at a smaller value of $\tau
$, this suggests that as $n$ increase, $\tau$ can decrease. As such,
spectral clustering should be able to correctly cluster the nodes in a
Stochastic Blockmodel graph when the minimum expected degree does not
grow linearly with the number of nodes in the graph.

Lemma \ref{frobeniusnormtheorem}, Theorem \ref{eigenconvergence}, and
Theorem \ref{growTheorem} all require the minimum expected degree to
grow at the same rate as $n$ (ignoring $\log n$ terms). Although the
strict assumption is inappropriate for large networks, this simulation
demonstrates (1) that spectral clustering works for smaller networks
and (2) that the asymptotic theory presented earlier in the paper can
be a guide to smaller networks. In these networks, it is not as
unreasonable that each node would be connected to a significant
proportion of the other nodes.\looseness=1

\section{Empirical edge density} \label{sec5}
In several networks, there is a natural or canonical notion of what an
edge represents. In an online social network, friendship is the
canonical notion of an edge. With this canonical notion, the edges in
most empirical networks are not dense enough to suggest the asymptotic
framework assumed in Lemma \ref{frobeniusnormtheorem}, Theorems \ref
{eigenconvergence} and \ref{growTheorem}.

Although it is an area of future research to weaken the strong
assumption on the expected node degrees, there are potentially other
notions of similarity that can replace the canonical notion. Define the
canonical edge set $E_c$ to contain $(i,j)$ if nodes $i$ and $j$ are
connected with a canonical edge. One possible extension of $E_c$ is
%
%
\begin{equation} \label{friendoffriend}
E_{ff} = \{(i,j) \dvtx\mbox{if $(i,k) \in E_c$ and $(k,j) \in E_c$
for some $k$}\}.
\end{equation}
In words, $(i,j) \in E_{ff}$ if $i$ and $j$ are friends of friends.

%
\begin{table}
\caption{This table describes five basic characteristics of the
Facebook social network within five universities. In the table below,
$\overline{\mathrm{deg}}{}^c$ is the average node degree using the canonical edges
of friendship and $\overline{\mathrm{deg}}{}^{ff}$ is the average node degree
using the ``friends-of-friends'' edges as defined with (\protect\ref
{friendoffriend}). The statistics $T_c$ and $T_{ff}$ [defined in
(\protect\ref{tc}) and (\protect\ref{tff})] are equal to the percent of nodes
that are connected to more than 10\% of the nodes in the graph. The
table below shows that the network is much more connected when using
edges defined by ``friends-of-friends.'' All numbers are rounded to the
nearest integer}\label{degrees}
\begin{tabular*}{\tablewidth}{@{\extracolsep{4in minus 4in}}ld{3.0}cccc@{}}
\hline
\textbf{School} & \multicolumn{1}{c}{\textbf{Caltech}} & \multicolumn{1}{c}{\textbf{Princeton}}
& \multicolumn{1}{c}{\textbf{Georgetown}} & \multicolumn{1}{c}{\textbf{Oklahoma}} & \multicolumn{1}{c@{}}{\textbf{UNC}}\\
\hline
$n$ & 769 & 6\mbox{,}596 &9\mbox{,}414 & 17\mbox{,}425 & 18\mbox{,}163 \\
$\overline{\mathrm{deg}}{}^c$ & 43 & \hphantom{00,}89 & \hphantom{00,}90 & \hphantom{00,}102 & \hphantom{000,}84\\
$\overline{\mathrm{deg}}{}^{ff}$ & 487 & 2\mbox{,}663 & 3\mbox{,}320 & \hphantom{0}5\mbox{,}420& \hphantom{0}5\mbox{,}242 \\
$T_c$ & 16 & \hphantom{000,}0& \hphantom{000,}0& \hphantom{0000,}0 & \hphantom{0000,}0\\
$T_{ff}$ & 94 & \hphantom{00,}88 & \hphantom{00,}87 & \hphantom{000,}81& \hphantom{000,}79 \\
\hline
\end{tabular*}
\end{table}

Table \ref{degrees} investigates the edge density of five empirical
network defined using both $E_c$ and $E_{ff}$. These five networks come
from the Facebook networks of five universities: California Institute
of Technology (Caltech), Princeton University, Georgetown University,
University of Oklahoma, University of North Carolina at Chapel Hill
(UNC). \citet{traud2008community} made these data sets publicly
available and investigated the community structure in them.

Let $W^c$ denote the adjacency matrix constructed from $E_c$. Let
$W^{ff}$ denote the adjacency matrix constructed from $E_{ff}$. Let
$\mathrm{deg}^c \in\R^n$ and $\mathrm{deg}^{ff} \in\R^n$ denote the degree sequences of
the nodes with respect to the two edge sets~$E_c$ and $E_{ff}$. That
is, $\mathrm{deg}_i^{ff} = \sum_j W_{ij}^{ff}$. Similarly\vspace*{1pt} for $\mathrm{deg}^c$. Define
%
%
\begin{eqnarray}
\label{dcbar}
\overline{\mathrm{deg}}{}^c & = & \frac{1}{n} \sum_i \mathrm{deg}_i^c,\\
\label{dffbar}
\overline{\mathrm{deg}}{}^{ff} & = & \frac{1}{n} \sum_i \mathrm{deg}_i^{ff},\\
\label{tc}
T_c &=& \frac{100 \%}{n} \sum_i \mathbf{1}\{\mathrm{deg}_i^c > n/10\}, \\
\label{tff}
T_{ff} &=& \frac{100 \%}{n} \sum_i \mathbf{1}\{\mathrm{deg}_i^{ff} > n/10\}.
\end{eqnarray}
The first two quantities are equal to the average node degrees. The
last two quantities are the percent of nodes connected to more that
10\% of the nodes in the network.

Table \ref{degrees} demonstrates how the edge density increases after
replacing $E_c$ with~$E_{ff}$. The statistics $T_c$ and $T_{ff}$, in
the last two lines of the table, can be used to gauge the suitability
of the assumption $\tau^2 > 2/\log n$ in the theorems above. Recall
that $\tau$ is the minimum expected degree divided by $n$. So, for
example, if $T_{ff} = 1$, then it is reasonable to expect that $\tau>
1/10$. Because there are some nodes that have a very small degree,
$T_c$ and $T_{ff}$ look at the proportion of nodes that are well connected.

It is an empirical observation that graphs have sparse degrees. This
suggests that the assumption $\tau^2 > 2 / \log n$ in Lemma \ref
{frobeniusnormtheorem}, Theorem \ref{eigenconvergence} and Theorem
\ref
{growTheorem} is not satisfied in practice. Table \ref{degrees}
demonstrates that by using an alternative notion of adjacency or
connected, the network can become much more connected.

\section{Discussion} \label{discussion}

The goal of this paper is to bring statistical rigor to the study of
community detection by assessing how well spectral clustering can
estimate the clusters in the Stochastic Blockmodel. The Stochastic
Blockmodel is easily amenable to the analysis of clustering algorithms
because of its simplicity and well-defined communities. The fact that
spectral clustering performs well on the Stochastic Blockmodel is
encouraging. However, because the Stochastic Blockmodel fails to
represent fundamental features that most empirical networks display,
this result should only be considered a first step.

This paper has two main results. The first main result, Theorem \ref
{eigenconvergence}, proves that under the latent space model, the
eigenvectors of the empirical normalized graph Laplacian converge to
the eigenvectors of the population normalized graph Laplacian---so long
as (1) the minimum expected degree grows fast enough and (2) the
eigengap that separates the leading eigenvalues from the smaller
eigenvalues does not shrink too quickly. This theorem has consequences
in addition to those related to spectral clustering.

Visualization is an important tool for social networks analysts
[\citet{liotta-graph}, \citet{freeman2000visualizing},
\citet{wasserman1994social}]. However,
there is little statistical understanding of these techniques under
stochastic models. Two visualization techniques, factor analysis and
multidimensional scaling, have variations that utilize the eigenvectors
of the graph Laplacian. Similar approaches were suggested for social
networks as far back as the 1950s [\citet{bock1952factors},
\citet{breiger1975algorithm}].
\citet{koren2005drawing} suggests visualizing the graph using the
eigenvectors of the
unnormalized graph Laplacian. The analogous method for the normalized
graph Laplacian would use the $i$th row of $X$ as the coordinates
for the $i$th node. Theorem~\ref{eigenconvergence} shows that, under
the latent space model, this visualization is not much different than
visualizing the graph by instead replacing $X$ with $\popx$. If there
is structure in the latent space of a latent space model (e.g.,
the $z_1,\ldots, z_n$ form clusters) and this structure is represented
in the eigenvectors of the population normalized graph Laplacian, then
plotting the eigenvectors will potentially reveal this structure.

The Stochastic Blockmodel is a specific latent space model that
satisfies these conditions. It has well-defined clusters or blocks and
Lemma \ref{mu} shows that, under weak assumptions, the eigenvectors of
the population normalized graph Laplacian perfectly identify the block
structure. Theorem \ref{eigenconvergence} suggests that you could
discover this clustering structure by using the visualization technique
proposed by \citet{koren2005drawing}. The second main result,\vadjust{\eject} Theorem
\ref{growTheorem}, goes further to suggest just how many nodes you
might miscluster by running $k$-means on those points (this is spectral
clustering). Theorem \ref{growTheorem} proves that if (1) the minimum
expected degree grows fast enough and (2)~the smallest nonzero
eigenvalue of the population normalized graph Laplacian shrinks slowly
enough, then the proportion of nodes that are misclustered by spectral
clustering vanishes in the asymptote.

The asymptotic framework applied in Theorem \ref{growTheorem} allows
the number of blocks to grow with the number of nodes; this is the
first such high-dimensional clustering result. Allowing the number of
clusters to grow is reasonable because as \citet
{leskovec2008statistical} noted, large networks do not necessarily have
large communities. In fact, in a wide range of empirical networks, the
tightest communities have a roughly constant size. Allowing the number
of blocks to grow with the number of nodes ensures the clusters do not
become too large.

There are two main limitations of our results that are highlighted in
the simulations in Section \ref{sec4}. First, Theorem \ref
{growTheorem} does not show that spectral clustering is consistent
under the Stochastic Blockmodel; it only gives a~bound on the number of
misclassified nodes. Improving this bound is an area for future
research. The second shortcoming is that Lemma~\ref{frobeniusnormtheorem}, Theorems \ref{eigenconvergence} and
\ref{growTheorem} all require the minimum expected degree to grow at the
same rate as $n$ (ignoring $\log n$ terms). In large empirical
networks, the canonical edges are not dense enough to suggest this type
of asymptotic framework. Section \ref{sec5} suggests alternative
definitions of edges that might increase the edge density. That said,
studying spectral clustering under more realistic degree distributions
is an area for future research.

\begin{appendix}

\vspace*{-2pt}\section{\texorpdfstring{Proof of Theorem \lowercase{\protect\ref{frobeniusnormtheorem}}}{Proof of Theorem 2.1}}\vspace*{-2pt}
\label{latentSpaceAppendix}

First, a proof of Lemma \ref{square}.
\begin{pf*}{Proof of Lemma \ref{square}}
By eigendecomposition, $M = \sum_{i=1}^n \lambda_i u_i u_i^T$ whe\-re
$u_1,\ldots, u_n$ are orthonormal and eigenvectors of $M$. So,
\[
MM = \Biggl( \sum_{i=1}^n \lambda_i u_i u_i^T \Biggr) \Biggl( \sum
_{i=1}^n \lambda_i u_i u_i^T \Biggr) = \sum_{i=1}^n \lambda_i^2 u_i u_i^T.
\]
Right multiplying by any $u_i$ yields $MM u_i = \lambda^2 u_i$. This
proves one direction of part one in the lemma, if $\lambda$ is an
eigenvalue of $M$, then $\lambda^2$ is an eigenvalue of $MM$. It also
proves part two of the lemma, all eigenvectors of $M$ are also
eigenvectors of $MM$.

To see that if $\lambda^2$ is an eigenvalue of $MM$, then $ \lambda$ or
$-\lambda$ is an eigenvalue of~$M$, notice that both $M$ and $MM$ have
exactly $n$ eigenvalues (counting multiplicities) because both matrices
are real and symmetric. So, the previous paragraph specifies $n$
eigenvalues of $MM$ by squaring the eigenvalues of~$M$. Because $MM$
has exactly $n$ eigenvalues, there are no other
eigenvalues.\looseness=-1

The rest of the proof is devoted to part three of the lemma. Let $MM v
= \lambda^2 v$. By eigenvalue decomposition, $M = \sum_i \lambda_i u_i
u_i^T$ and because $u_1,\ldots, u_n$ are orthonormal ($M$ is real and
symmetric) there exists $\alpha_1,\ldots, \alpha_n$ such that $v =
\sum
_i \alpha_i u_i$.
\begin{eqnarray*}
\lambda^2 \sum_i \alpha_i u_i &=& \lambda^2 v = MM v = M\biggl( \sum_i
\lambda_i u_i u_i^T v\biggr) = M\biggl( \sum_i \lambda_i \alpha_i u_i
\biggr) \\
&=& \sum_i \lambda_i \alpha_i M u_i = \sum_i \lambda_i^2 \alpha_i u_i.
\end{eqnarray*}
By the orthogonality of the $u_i$'s, it follows that $\lambda^2 \alpha
_i = \lambda_i^2 \alpha_i$ for all $i$. So, if $\lambda_i^2 \ne
\lambda
^2$, then $\alpha_i=0$.
\end{pf*}

For $i = 1,\ldots, n$, define $c_i = \popd_{ii} / n$ and
$\tau= \min_{i = 1,\ldots, n} c_i$.
\begin{lemma} \label{mainThm} If $n^{1/2} / \log n > 2$,
\[
\bP\biggl(\|L L - \popl\popl\|_F \ge\frac{32 \sqrt{2} \log n}{ \tau^{2}
n^{1/2 }} \biggr) \le4 n^{2 - 2 \tau^2 \log n}.
\]
\end{lemma}

The main complication of the proof of Lemma \ref{mainThm} is
controlling the dependencies between the elements of $L L$. We do this
with an intermediate step that uses the matrix
\[
\tilde L = \popd^{-1/2}W\popd^{-1/2}
\]
and two sets $ \Gamma$ and $\Lambda$. $\Gamma$ constrains the matrix
$D$, while $\Lambda$ constrains the matrix $W \popd^{-1} W$. These sets
will be defined in the proof. To ease the notation, define
\[
\bP_{\Gamma\Lambda}(B) = \bP(B \cap\Gamma\cap\Lambda),
\]
where $B$ is some event.
\begin{pf*}{Proof of Lemma \ref{mainThm}}
This proof shows that under the sets $\Gamma$ and $\Lambda$ the
probability of the norm exceeding $32 \sqrt{2}\log(n) \tau^{-2}
n^{- 1/2 } $ is exactly zero for large enough $n$ and that the
probability of $\Gamma$ or $\Lambda$ not happening is exponentially
small. To ease notation, define $a = 32 \sqrt{2} \log(n) \tau^{-2}
n^{- 1/2 }$.

The diagonal terms behave differently than the off diagonal terms. So,
break them apart:
\begin{eqnarray*}
&&\bP(\|L L - \popl\popl\|_F \ge a ) \\
&&\qquad\le \bP_{\Gamma\Lambda}(\|L
L -\popl\popl\|_F \ge a ) + \bP\bigl( (\Gamma\cap\Lambda)^c \bigr)\\
&&\qquad=\bP_{\Gamma\Lambda}\biggl( \sum_{i,j} [L L - \popl\popl]_{ij}^2 \ge
a^2 \biggr) + \bP\bigl( (\Gamma\cap\Lambda)^c \bigr)\\
&&\qquad \le\bP_{\Gamma\Lambda}\biggl( \sum_{i \ne j} [L L - \popl\popl
]_{ij}^2 \ge a^2/2 \biggr) \\
&&\qquad\quad{} + \bP_{\Gamma\Lambda}\biggl( \sum_{i} [L L - \popl\popl
]_{ii}^2 \ge a^2/2 \biggr) \\
&&\qquad\quad{} + \bP\bigl( (\Gamma\cap\Lambda)^c \bigr).
\end{eqnarray*}
First, address the sum over the off diagonal terms:
%
%
\begin{eqnarray}
&& \bP_{\Gamma\Lambda}\biggl( \sum_{i \ne j} [L L - \popl\popl]_{ij}^2
\ge a^2/2 \biggr) \nonumber\\
\label{n2terms}
&&\qquad \le \bP_{\Gamma\Lambda}\biggl( \bigcup_{i\ne j} \biggl\{ [L L - \popl\popl
]_{ij}^2 \ge\frac{a^2}{ 2n^{2}} \biggr\} \biggr) \\
&&\qquad \le \sum_{i\ne j} \bP_{\Gamma\Lambda}\biggl( |L L - \popl\popl
|_{ij} \ge\frac{a }{\sqrt{2}n} \biggr) \nonumber\\
&&\qquad \le \sum_{i\ne j} \bP_{\Gamma\Lambda}\biggl( |L L - \tilde L \tilde
L|_{ij} + |\tilde L \tilde L - \popl\popl|_{ij} \ge\frac{a }{\sqrt
{2}n} \biggr) \nonumber\\
&&\qquad \le \sum_{i\ne j} \biggl[ \bP_{\Gamma\Lambda}\biggl( |L L - \tilde L
\tilde L|_{ij}\ge\frac{a}{\sqrt{8}n}\biggr) \nonumber\\
\label{offD}
&&\hphantom{\sum_{i\ne j}\biggl[}\qquad\quad{} + \bP_{\Gamma\Lambda}\biggl(|\tilde L \tilde L -
\popl\popl|_{ij} \ge\frac{a }{ \sqrt{8}n}\biggr) \biggr].
\end{eqnarray}
The sum over the diagonal terms is similar,
\begin{eqnarray*}
&&\bP_{\Gamma\Lambda}\biggl( \sum_{i} [L L - \popl\popl]_{ii}^2 \ge a^2/2
\biggr) \\
&&\qquad\le \sum_{i} \biggl[\bP_{\Gamma\Lambda}\biggl( |L L - \tilde L
\tilde L|_{ii}\ge\frac{ a }{ \sqrt{8n}}\biggr)
+ \bP_{\Gamma\Lambda}\biggl(|\tilde L \tilde L - \popl\popl
|_{ii} \ge\frac{a }{ \sqrt{8n}}\biggr)\biggr]
\end{eqnarray*}
with one key difference. In (\ref{n2terms}), the union bound
address nearly $n^2$ terms. This yields the $1/n^2$ term in line (\ref
{n2terms}). After taking the square root, each term has a lower bound
with a factor of $1/n$. However, because there are only~$n$ terms on
the diagonal, after taking\vspace*{1pt} the square root in the last equation above,
the lower bound has a factor of $1/\sqrt{n}$.

To constrain the terms $|\tilde L \tilde L - \popl\popl|_{ij}$ for
$i=j$ and $i \ne j$, define
\[
\Lambda= \bigcap_{i,j} \biggl\{ \biggl| \sum_k (W_{ik}W_{jk} -
p_{ijk})/c_k \biggr| < n^{1/2} \log n \biggr\},
\]
where
\[
p_{ijk} = \cases{
p_{ik}p_{jk}, &\quad if $i \ne j$, \cr
p_{ik}, &\quad if $i=j$,}
\]
for $p_{ij} = \popw_{ij}$. We now show that for large enough $n$ and
any $i \ne j$,
%
%
\begin{eqnarray}
\label{term2inej}
\bP_\Lambda\biggl(|\tilde L \tilde L - \popl\popl|_{ij} \ge\frac{ a }{
\sqrt{8}n}\biggr) &=& 0, \\
\label{term2iej}
\bP_\Lambda\biggl(|\tilde L \tilde L - \popl\popl|_{ii} \ge\frac{a }{
\sqrt{8n}}\biggr) &=& 0.
\end{eqnarray}
To see (\ref{term2inej}), expand the left-hand side of the
inequality for $i \ne j$,
\begin{eqnarray*}
|\tilde L \tilde L - \popl\popl|_{ij} &=& \frac{1}{(\popd_{ii}
\popd
_{jj})^{1/2}} \biggl| \sum_k (W_{ik}W_{jk} - p_{ik}p_{jk})/\popd_{kk}
\biggr| \\
&=& \frac{1}{n^2\sqrt{c_ic_j}} \biggl| \sum_k (W_{ik}W_{jk} -
p_{ik}p_{jk})/c_k \biggr|.
\end{eqnarray*}
This is bounded on $\Lambda$, yielding
\[
|\tilde L \tilde L - \popl\popl|_{ij} < \frac{\log n}{ \tau n^{3/2}}
\le\frac{32 \sqrt{2} \log n}{ \sqrt{8} \tau^{2} n^{3/2 }} = \frac
{a}{\sqrt{8} n}.
\]
So, (\ref{term2inej}) holds for $i \ne j$. Equation (\ref{term2iej})
is different because $W_{ik}^2 = W_{ik}$. As a result, the
diagonal of $\tilde L \tilde L$ is a biased estimator of the diagonal
of $\popl\popl$.
%
%
\begin{eqnarray} \label{onD2terms}\quad
|\tilde L \tilde L - \popl\popl|_{ii}
&=& \biggl| \sum_k \frac{W_{ik}^2 -
p_{ik}^2}{\popd_{ii}\popd_{kk}} \biggr|
= \biggl| \sum_k \frac{W_{ik} - p_{ik}^2}{\popd_{ii}\popd_{kk}}
\biggr|\nonumber\\
&\le& \biggl| \sum_k \frac{W_{ik} - p_{ik}}{\popd_{ii}\popd_{kk}} \biggr| +
\biggl| \sum_k \frac{p_{ik} - p_{ik}^2}{\popd_{ii}\popd_{kk}} \biggr|
\\
&=&  \frac{1}{c_i n^2} \biggl( \biggl| \sum_k (W_{ik} - p_{ik})/c_k \biggr|
+ \biggl| \sum_k (p_{ik} - p_{ik}^2)/c_k \biggr| \biggr).\nonumber
\end{eqnarray}
Similar to the $i \ne j$ case, the first term is bounded by $ \log(n)
\tau^{-1} n^{- 3/2}$ on $\Lambda$. The second term is bounded by
$\tau
^{-2} n^{-1}$:
\[
\frac{1}{c_i n^2} \biggl| \sum_k (p_{ik} - p_{ik}^2)/c_k \biggr|  \le
\frac{1}{c_i n^2} \biggl| \sum_k 1/\tau\biggr|
\le\frac{1}{\tau^2 n}.
\]
Substituting the value of $a$ in reveals that on the set $\Lambda$,
both terms in (\ref{onD2terms}) are bounded by $a(2\sqrt{8n})^{-1}$.
So, their their sum is bounded by $a(\sqrt{8n})^{-1}$, satisfying
(\ref{term2iej}).

This next part addresses the difference between $LL$ and $\tilde L
\tilde L$, showing that for large enough $n$, any $i \ne j$, and some
set $\Gamma$,
\begin{eqnarray*}
\bP_{\Gamma\Lambda}\biggl( |L L - \tilde L \tilde L|_{ij}\ge\frac{a}{
\sqrt
{8}n}\biggr) &=& 0, \\
\bP_{\Gamma\Lambda}\biggl( |L L - \tilde L \tilde L|_{ii}\ge\frac{a}{
\sqrt
{8n}}\biggr) & = & 0.
\end{eqnarray*}
It is enough to show that for any $i$ and $j$,
%
%
\begin{equation} \label{bound2}
\bP_{\Gamma\Lambda}\biggl( |L L - \tilde L \tilde L|_{ij}\ge\frac{a }{
\sqrt
{8}n}\biggr) =0.
\end{equation}

For $b(n) = \log(n) n^{-1/2}$, define $u(n) = 1 + b(n), l(n) = 1-
b(n)$. With these define the following sets:
\begin{eqnarray*}
\Gamma&= &\bigcap_i \{D_{ii} \in\popd_{ii} [l(n), u(n)]\},
\\
\Gamma(1) &= &\bigcap_i \biggl\{\frac{1}{D_{ii}} \in\frac{1}{\popd
_{ii}} [u(n)^{-1}, l(n)^{-1}]\biggr\}, \\
\Gamma(2) &=&\bigcap_{i,j} \biggl\{\frac{1}{(D_{ii}D_{jj})^{1/2}} \in
\frac{1}{(\popd_{ii}\popd_{jj})^{1/2}} [u(n)^{-1}, l(n)^{-1}
]\biggr\}, \\
\Gamma(3) &=&\bigcap_{i,j,k} \biggl\{\frac
{1}{D_{kk}(D_{ii}D_{jj})^{1/2}} \in\frac{[u(n)^{-2},
l(n)^{-2}]}{\popd_{kk}(\popd_{ii}\popd_{jj})^{1/2}} \biggr\}.
\end{eqnarray*}

Notice that $\Gamma\subseteq\Gamma(1) \subseteq\Gamma(2)$ and
$\Gamma\subseteq\Gamma(3)$. Define another set:
\[
\Gamma(4) = \bigcap_{i,j,k} \biggl\{\frac{1}{D_{kk}(D_{ii}D_{jj})^{1/2}}
\in\frac{[1 - 16b(n),1 + 16b(n) ]}{\popd_{kk}(\popd
_{ii}\popd_{jj})^{1/2}} \biggr\}.
\]
The next steps show that this set contains $\Gamma$. It is sufficient
to show $\Gamma(3) \subset\Gamma(4)$. This is true because
\begin{eqnarray*}
\frac{1}{u(n)^2} &=& \frac{1}{(1+b(n))^2} = \frac{b(n)^{-2}}{(b(n)^{-1}
+ 1)^2} > \frac{b(n)^{-2} - 1 }{(b(n)^{-1} + 1)^2} \\
&=& \frac{b(n)^{-1} - 1}{b(n)^{-1} + 1} = 1 - \frac{2}{b(n)^{-1} + 1} >
1 - 16 b(n).
\end{eqnarray*}
The $16$ in the last bound is larger than it needs to be so that the
upper and lower bounds in $\Gamma(4)$ are symmetric. For the other direction,
\begin{eqnarray*}
\frac{1}{l(n)^2} &=& \frac{1}{(1- b(n))^2} = \frac
{b(n)^{-2}}{(b(n)^{-1} - 1)^2} = \biggl( 1 + \frac{1}{b(n)^{-1}- 1}
\biggr)^2 \\
&=& 1 + \frac{2}{b(n)^{-1}- 1} + \frac{1}{(b(n)^{-1}- 1)^2} .
\end{eqnarray*}

We now need to bound the last two elements here. We are assuming,
$\sqrt{n} / \log n > 2$. Equivalently, $1- b(n) > 1/2$. So,
we have both of the following:
\[
\frac{1}{(b(n)^{-1}- 1)^2} < \frac{2}{b(n)^{-1}- 1}
\quad\mbox{and}\quad \frac{2}{b(n)^{-1}- 1} = \frac{2b(n)}{1-
b(n)} < 8 b(n).
\]
Putting these together,
\[
\frac{1}{l(n)^2} < 1 + 16 b(n).
\]
This shows that $\Gamma\subset\Gamma(4)$. Now, under the set $\Gamma
$, and thus $\Gamma(4)$,
\begin{eqnarray*}
|L L - \tilde L \tilde L|_{ij} & = & \biggl| \sum_k \biggl( \frac
{W_{ik}W_{jk}}{D_{kk}(D_{ii}D_{jj})^{1/2}} - \frac{W_{ik}W_{jk}}{\popd
_{kk}(\popd_{ii}\popd_{jj})^{1/2}} \biggr)\biggr| \\[-2pt]
& \le& \sum_k \biggl| \frac{1}{D_{kk}(D_{ii}D_{jj})^{1/2}} - \frac
{1}{\popd_{kk}(\popd_{ii}\popd_{jj})^{1/2}}\biggr| \\[-2pt]
& \le& \sum_k \biggl|\frac{16 b(n)}{\popd_{kk}(\popd_{ii}\popd
_{jj})^{1/2}} \biggr| \\[-2pt]
& \le& \sum_k \frac{16 b(n)}{\tau^2 n^2}
\le \frac{16 b(n)}{\tau^2 n}.
\end{eqnarray*}
This is equal to $a (\sqrt{8}n)^{-1} $, showing (\ref{term2iej})
holds for all $i$ and $j$.

The remaining step is to bound $\bP((\Gamma\cap\Lambda)^c)$. Using
the union bound, this is less than or equal to $\bP(\Gamma^c) + \bP
(\Lambda^c)$:
\begin{eqnarray*}
\bP(\Gamma^c) &=& \bP\biggl(\bigcup_i \{D_{ii} \notin\popd_{ii}
[1-b(n), 1 + b(n)]\}\biggr)\\[-2pt]
&\le& \sum_i \bP\bigl( \{D_{ii} \notin\popd_{ii} [1-b(n), 1 +
b(n)]\}\bigr)\\[-2pt]
&<& \sum_i 2\exp\biggl(-2 \biggl(\frac{\popd_{ii} \log n}{\sqrt{n}}
\biggr)^2 \frac{1}{n}\biggr) \\[-2pt]
& \le& 2 n \exp(-2 \tau^2 (\log n)^2) \\[-2pt]
& = & 2 n^{1 - 2 \tau^2 \log n},
\end{eqnarray*}
where the second to last inequality is by Hoeffding's inequality. The
next inequality is Hoeffding's:
\begin{eqnarray*}
\bP(\Lambda^c) &=& \bP\biggl(\bigcup_{i,j} \biggl\{ \biggl| \sum_k
(W_{ik}W_{jk} - p_{ijk})/c_k \biggr| > n^{1/2} \log n \biggr\}\biggr) \\[-2pt]
&=& \sum_{i, j} \bP\biggl( \biggl| \sum_k (W_{ik}W_{jk} - p_{ijk})/c_k
\biggr| > n^{1/2} \log n \biggr) \\[-2pt]
&<& \sum_{i, j} 2 \exp\biggl(- 2 n (\log n)^2 \Big/ \sum_k 1/c_k^2 \biggr)\\[-2pt]
& \le& \sum_{i,j} 2 \exp(-2 (\log n)^2 \tau^2 )\\[-2pt]
& \le& 2 n^2 \exp(-2 (\log n)^2 \tau^2 )\\[-2pt]
& \le& 2 n^{2 - 2\tau^2 \log n}.
\end{eqnarray*}
Because $W$ is symmetric, the independence of the $W_{ik}W_{jk}$ across
$k$ is not obvious. However, because $W_{ii} = W_{jj} = 0$, they are
independent across $k$.

Putting the pieces together,
\begin{eqnarray*}
&&\bP\biggl(\|L L - \popl\popl\|_F \ge\frac{32 \sqrt{2} \log n}{ \tau
^{2} n^{1/2 }} \biggr) \\
&&\qquad\le\bP_{\Gamma\Lambda}\biggl(\|L L - \popl\popl\|_F \ge\frac{32
\sqrt{2}
\log n}{ \tau^{2} n^{1/2 }} \biggr) \\
&&\qquad\quad{}+ \bP\bigl( (\Gamma\cap\Lambda)^c
\bigr)\\
&&\qquad <  0 + 2 n^{1 - 2 \tau^2 \log n}+ 2 n^{2 - 2\tau^2 \log n}\\
&&\qquad\le 4 n^{2 - 2 \tau^2 \log n}.
\end{eqnarray*}
\upqed\end{pf*}

The following proves Theorem \ref{frobeniusnormtheorem}.
\begin{pf*}{Proof of Theorem \ref{frobeniusnormtheorem}}
Adding the $n$ super- and subscripts to Lem\-ma~\ref{mainThm}, it states
that if $n^{1/2} / \log n > 2$, then
\[
\bP\biggl(\|L L - \popl\popl\|_F \ge\frac{c \log n}{ \tau^{2} n^{1/2
}} \biggr) < 4 n^{2 - 2 \tau^2 \log n}
\]
for $c = 32 \sqrt{2}$. By assumption, for all $n > N$, $\tau_n^2 \log n
> 2$. This implies that $2 - 2 \tau_n^2 \log n < -2$ for all $n > N$.
Rearranging and summing over $n$, for any fixed $ \varepsilon>0$,
\begin{eqnarray*}
\sum_{n=1}^\infty\bP\biggl( \frac{\|\Ln\Ln- \popln\popln\|_F}{c \tau
_n^{-2} \log(n) n^{-1/2 } / \varepsilon} \ge\varepsilon\biggr) &\le&
N+4 \sum_{n=N+1}^\infty n^{2 - 2 \tau_n^2 \log n} \\
&\le& N+4 \sum_{n=N+1}^\infty n^{-2},
\end{eqnarray*}
which is a summable sequence. By the Borel--Cantelli theorem,
\[
\bigl\| \Ln\Ln- \popln\popln\bigr\|_F = o( \tau_n^{-2} \log(n) n^{- 1/2})
\qquad\mbox{a.s.}
\]
\upqed\end{pf*}

\section{Davis--Kahan theorem} \label{dkappendix}

The statement of the theorem below and the preceding explanation come
largely from \citet{vonluxburg2007tsc}. For a more detailed
account of
the Davis--Kahan theorem, see \citet{gw1990matrix}.

To avoid the issues associated with multiple eigenvalues, this
theorem's original statement is instead about the subspace formed by
the eigenvectors. For a distance between subspaces, the theorem uses
``canonical angles,'' which are also known as ``principal angles.'' Given
two matrices $M_1$ and $M_2$ both in $\R^{n\times p}$ with orthonormal
columns, the singular values ($\sigma_1,\ldots, \sigma_p$) of~$M_1'M_2$
are the cosines of the principal angles ($\cos\Theta_1,\ldots, \cos
\Theta_p$) between the column space of $M_1$ and the column space of
$M_2$. Define $\sin\Theta(M_1, M_2)$ to be a diagonal matrix
containing the sine of the principal angles of $M_1'M_2$ and define
%
%
\begin{equation}\label{dist}
d(M_1, M_2) = \| {\sin\Theta}(M_1, M_2) \|_F,
\end{equation}
which can be expressed as $(p - \sum_{j=1}^p \sigma_j^2)^{1/2}$ by
using the identity $\sin^2 \theta= 1 - \cos^2 \theta$.
\begin{prop}[(Davis--Kahan)] Let $S \subset\R$ be an interval.
Denote~$\mathcal{X}$ as an orthonormal matrix whose column space is
equal to the eigenspace of $\popl\popl$ corresponding to the
eigenvalues in $\lambda_S(\popl\popl)$ [more formally, the column
space of $\mathcal{X}$ is the image of the spectral projection of
$\popl\popl$ induced by $\lambda_S(\popl\popl)$]. Denote by $X$ the
analogous quantity for $LL$. Define the distance between $S$ and the
spectrum of $\popl\popl$ outside of $S$ as
\[
\delta= \min\{ |\lambda- s|; \lambda\mbox{ eigenvalue of } \popl
\popl, \lambda\notin S, s \in S\}.
\]
Then the distance $d(\mathcal{X}, X) = \| {\sin\Theta}(\mathcal{X},
X) \|
_F$ between the column spaces of~$\mathcal{X}$ and $X$ is bounded by
\[
d(X, \mathcal{X}) \le\frac{ \|LL - \popl\popl\|_F}{\delta}.
\]
\end{prop}

In the theorem, $\popl\popl$ and $LL$ can be replaced by any two
symmetric matrices. The rest of this section converts the bound on
$d(X, \mathcal{X})$ to a bound on $\|X - \mathcal{X} O \|_F$, where $O$
is some orthonormal rotation. For this, we will make an additional
assumption that $\mathcal{X}$ and $X$ have the same dimension. Assume
there exists $S \subset\R$ containing $k$ eigenvalues of $\popl\popl$
and $k$ eigenvalues of $LL$, but containing no other eigenvalues of
either matrix. Because $L L$ and $\popl\popl$ are symmetric, its
eigenvectors can be defined to be orthonormal. Let the columns of
$\popx\in\R^{n \times k}$ be $k$ orthonormal eigenvectors of $\popl
\popl$ corresponding to the $k$ eigenvalues contained in $S$. Let the
columns of $X\in\R^{n \times k}$ be~$k$ orthonormal eigenvectors of
$LL$ corresponding to the $k$ eigenvalues contained in $S$.
By singular value decomposition, there exist orthonormal~matri\-ces~$U,V$ and diagonal matrix $\Sigma$ such that
$\popx^T X = U \Sigma V^T$. The singular values, $\sigma_1,\ldots,
\sigma_k$, down the diagonal of $\Sigma$ are the cosines of the
principal angles between the columns space of $X$ and the column space
of $\popx$.

Although the Davis--Kahan theorem is a statement regarding the principal
angles, a few lines of algebra show that it can be extended to a bound
on the Frobenius norm between\vadjust{\goodbreak} the matrix $X$ and $\popx U V^T$, where
the matrix~$U V^T$ is an orthonormal rotation:
\begin{eqnarray*}
\frac{1}{2}\| X - \popx U V^T \|_F^2 &=& \frac{1}{2} \operatorname{trace}\bigl((X - \popx U
V^T )^T(X - \popx U V^T )\bigr) \\
&=& \frac{1}{2} \operatorname{trace}(VU^T\popx^T\popx U V^T + X^T X - 2VU^T\popx^TX )
\\
&=& \frac{1}{2}\bigl( k + k - 2\operatorname{trace}(VU^T\popx^TX )\bigr) \\
&\le& \frac{\|L L - \popl\popl\|_F^2}{\delta^2},
\end{eqnarray*}
where the last inequality is explained below. It follows from a
property of the trace, the fact that the singular values are in
$[0,1]$, the trigonometric identity $\cos^2 \theta= 1 - \sin^2
\theta$
and the Davis--Kahan theorem:
\begin{eqnarray*}
\operatorname{trace}( V U^T \popx^T X ) &=& \sum_{i=1}^k \sigma_i \ge\sum_{i=1}^k
(\cos\Theta_i)^2 =\sum_{i=1}^k 1 - (\sin\Theta_i)^2 \\
&=& k - (d(X,\popx))^2 \ge k - \frac{\|LL - \popl\popl\|
_F^2}{\delta^2}.
\end{eqnarray*}
This shows that the Davis--Kahan theorem can instead be thought of as
a~bounding $\| \popx U V^T - X \|_F^2$ instead of $d(\popx, X)$. The
matrix $O$ in Theorem~\ref{DavisKahan} is equal to $UV^T$. In this way,
it is dependent on $X$ and $\mathcal{X}$.

\vspace*{3pt}\section{\texorpdfstring{Proof of Theorem \lowercase{\protect\ref{eigenconvergence}}}{Proof of Theorem
2.2}}\vspace*{3pt}
\label{proofeigenconvergence}

By Lemma \ref{square}, the column vectors of $X_n$ are eigenvectors of
$\Ln\Ln$ corresponding to all the eigenvalues in $\lambda_{S_n}(\Ln
\Ln
)$. For the application of the Davis--Kahan theorem, this means that the
column space of $X_n$ is the image of the spectral projection of $\Ln
\Ln$ induced by $\lambda_{S_n}(\Ln\Ln)$, similarly for the column
vectors of $\popx_n$, the matrix $\popln\popln$ and the set
$\lambda
_{S_n}(\popln\popln)$.

Recall that $\bar\lambda_1^{(n)}\ge\cdots\ge\bar\lambda_n^{(n)}$
are defined to be the eigenvalues of symmetric matrix $\popln\popln$
and $ \lambda_1^{(n)}\ge\cdots\ge\lambda_n^{(n)}$ are defined to be
the eigenvalues of symmetric matrix $\Ln\Ln$. From (\ref{evbound}),
\[
\max_i \bigl| \lambda^{(n)}_i - \bar\lambda^{(n)}_i\bigr| = o\biggl(\frac{\log
n}{\tau_n^2 n^{1/2}}\biggr).
\]
By assumption, $\tau_n^2 > 2/\log n$. So,
\[
\frac{\log n}{\tau_n^2 n^{1/2}} < \frac{(\log n)^2}{2 n^{1/2}} =
O(\min
\{\delta_n, \delta_n'\}),
\]
where the last step follows by assumption. Thus,
\[
\max_i \bigl| \lambda^{(n)}_i - \bar\lambda^{(n)}_i\bigr| = o(\min\{\delta_n,
\delta_n'\}).
\]
This means that, eventually, $\lambda_i^{(n)} \in S_n$ if and only if
$\bar\lambda^{(n)}_i \in S_n$. Thus, the number of elements in
$\lambda
_{S_n}(\popln\popln)$ is eventually equal to the number of elements in
$\lambda_{S_n}(\Ln\Ln)$ implying that $X_n$ and $\popx_n$ will
eventually have the same number of columns, $k_n = \mathscr{K}_n$.

Once $X_n$ and $\popx_n$ have the same number of columns, define
matrices~$U_n$ and $V_n$ with singular value decomposition: $\popx_n^T
X_n = U_n \Sigma_n V_n^T$. Define $O_n = U_n V_n^T$. The result follows
from the Davis--Kahan theorem and Theorem~\ref{frobeniusnormtheorem}:
\[
\| X_n - \popx_n O_n \|_F \le\frac{2 \|\Ln\Ln- \popln\popln\|
_F}{\delta_n} = o\biggl( \frac{\log n}{\delta_n \tau_n^{2} n^{ 1/2}}
\biggr) \qquad\mbox{a.s.}
\]

\section{Stochastic Blockmodel}
\label{sbmappendix}

%
\begin{pf*}{Proof of Lemma \ref{mu}}
First, construct the matrix $B_L \in R^{k \times k}$ such that $\popl
= Z B_L Z^T$. Define $D_B = \operatorname{diag}(BZ^T\mathbf{1}_n) \in\R^{k \times k}$
where $\mathbf{1}_n$ is a vector of ones in $\R^n$. For any $i,j$,
\[
\popl_{ij} = \frac{\popw_{ij}}{\sqrt{\popd_{ii} \popd_{jj}}} = z_i
D_B^{-1/2} B D_B^{-1/2} (z_j)^T.
\]
Define $B_L = D_B^{-1/2} B D_B^{-1/2}$. It follows that $\popl_{ij} =
(ZB_LZ^T)_{ij}$ and thus $\popl= ZB_LZ^T$.

Because $B$ is symmetric, so are $B_L$ and $(Z^TZ)^{1/2} B_L
(Z^TZ)^{1/2}$. Notice that
\[
\operatorname{det}((Z^TZ)^{1/2} B_L (Z^TZ)^{1/2}) = \operatorname{det}((Z^TZ)^{1/2}) \operatorname{det}( B_L )
\operatorname{det}((Z^TZ)^{1/2}) >0.
\]
By eigenvector decomposition, define $V \in R^{k \times k}$ and
diagonal matrix $\Lambda\in R^{k \times k}$ such that
%
%
\begin{equation} \label{Vdef}
(Z^TZ)^{1/2} B_L (Z^TZ)^{1/2} = V \Lambda V^T.
\end{equation}
Because the determinant of the left-hand side of (\ref{Vdef})
is greater than zero, none of the eigenvalues in $\Lambda$ are equal to
zero. Left multiply (\ref{Vdef}) by $Z(Z^TZ)^{-1/2}$ and right
multiply by $(Z^TZ)^{-1/2}Z^T$. This shows
%
%
\begin{equation}\label{Zmudef}
ZB_LZ^T = Z \mu\Lambda(Z \mu)^T,
\end{equation}
where $\mu= (Z^TZ)^{-1/2}V$. Notice that $(Z \mu)^T (Z \mu) = I_k$,
the $k \times k$ identity matrix. So, right multiplying (\ref{Zmudef})
by $Z\mu$ shows that the columns\vadjust{\eject} of $Z\mu$ are eigenvectors of
$ZB_LZ^T = \popl$ with the eigenvalues down the diagonal of $\Lambda$.
Equation (\ref{Zmudef}) shows that these are the only nonzero eigenvalues.

It remains to prove equivalence statement (\ref{zeq}). Notice
\[
\operatorname{det}(\mu) = \operatorname{det}((Z^TZ)^{-1/2}) \operatorname{det}(V) >0.
\]
So, $\mu^{-1}$ exists and statement (\ref{zeq}) follows.
\end{pf*}

The following is a proof of Lemma \ref{misclusteredLemma}.
\begin{pf*}{Proof of Lemma \ref{misclusteredLemma}}
The following statement is the essential ingredient to prove Lemma \ref
{misclusteredLemma}.
%
%
\begin{equation} \label{faraway}
z_i \ne z_j\qquad \mbox{then } \|z_i\mu- z_j \mu\|_2 \ge\sqrt{2/P}.
\end{equation}
The proof of statement (\ref{faraway}) requires the following
definition:
\[
\|\mu\|_m^2 = \min_{x\dvtx\|x\|_2 =1} \|x\mu\|_2^2.
\]
Notice that
\[
\|\mu\|_m^2 = \min_{x\dvtx\|x\|_2 =1} x\mu\mu^T x^T = \min
_{x\dvtx\|x\|_2
=1} x (Z^TZ)^{-1} x^T = 1/P.
\]
So,
\[
\|z_i\mu- z_j \mu\|_2 = \|(z_i - z_j) \mu\|_2 \ge\sqrt{2} \|\mu\|
_m =
\sqrt{2/P}.
\]
Proving statement (\ref{faraway}). The proof of Lemma \ref
{misclusteredLemma} follows:
\[
\|c_i O - z_j \mu\|_2 \ge\|z_i \mu- z_j \mu\|_2 - \|c_i O - z_i \mu
\|
_2 \ge\sqrt{\frac{2}{P}} - \frac{1}{2}\sqrt{\frac{2}{P}} = \frac
{1}{\sqrt{2P}}.\quad
\]
\upqed\end{pf*}
%
%
\begin{pf*}{Proof of Theorem \ref{growTheorem}}
Define $X \in R^{n \times k}$ to contain the eigenvectors of $L$
corresponding to the largest $k$ eigenvalues and define
\[
C =\mathop{\arg\min}_{M \in\mathscr{R}(n,k)} \|X - M\|_F^2,
\]
where $\mathscr{R}(n,k)$ is defined as:
\[
\mathscr{R}(n,k) = \{ M \in R^{n \times k} \dvtx M \mbox{ has no more
than $k$ unique rows}\}.
\]
Notice that
\[
\min_{M \in\mathscr{R}(n,k)} \|X - M\|_F^2 = \min_{\{ m_1,\ldots,
m_k\} \subset\R^k} \sum_i \min_g \|x_i - m_g\|_2^2.
\]
This shows that the $i$th row of $C$ is equal to $c_i$ as defined in
Definition \ref{cdef}. Because $Z\mu O \in\mathscr{R}(n,k)$, notice that
%
%
\begin{equation}\label{kmeansbound}
\|X - C \|_2 \le\|X - Z \mu O\|_2 .
\end{equation}
By the triangle inequality and inequality \ref{kmeansbound},
\[
\|C - Z \mu O\|_2 \le\|C - X\|_2 + \|X - Z \mu O\|_2 \le2 \|X - Z \mu
O\|_2.
\]\vadjust{\eject}

In the notation of Theorem \ref{eigenconvergence}, define $S_n =
[\lambda_{k_n}^2/2,2]$. Then, $\delta= \delta' = \lambda_{k_n}^2/2$.
By assumption, $n^{-1/2} (\log n)^2 = O(\lambda_{k_n}^2) = O(\min\{
\delta, \delta'\})$. This implies that the results from Theorem \ref
{eigenconvergence} hold. Putting the pieces together,
\begin{eqnarray*}
| \mathscr{M}| &\le&\sum_{i \in\mathscr{M}} 1 \le 2 P_n \sum_{i
\in
\mathscr{M}} \|c_i - z_i\mu O\|_2^2 \\
&\le& 2 P_n \|C - Z \mu O \|_F^2 \\
&\le& 8 P_n \|X - Z \mu O \|_F^2 \\
& =& o\biggl( \frac{P_n (\log n)^2}{n \lambda_{k_n}^4 \tau_n^{4} }
\biggr) \qquad\mbox{ a.s.}
\end{eqnarray*}
\upqed\end{pf*}

In the second example of Section \ref{sec3}, it was claimed that
\[
\lambda_k = \frac{1}{k(r/p) + 1}.
\]
The following is a proof of that statement.

Define $B \in\R^{k \times k}$ such that
\[
B = pI_{k} + r\mathbf{1}_{k}\mathbf{1}_{k}^T,
\]
where $I_{k} \in\R^{k \times k}$ is the identity matrix, $\mathbf
{1}_{k} \in\R
^{k}$ is a vector of ones, $r\in(0,1)$ and $p \in(0,1-r)$. Assume
that $p$ and $r$ are fixed and $k$ can grow with $n$. Let $Z \in\{0,1\}
^{n \times k}$ be such that $Z^T \mathbf{1}_n = s \mathbf{1}_k$. This
guarantees that
all $k$ groups have equal size $s$. The Stochastic Blockmodel in
the second example of Section \ref{sec3}
has the population adjacency matrix, $\popw=Z BZ^T$.

Define
\[
B_L = \frac{1}{nr + sp} (pI_{k} + r\mathbf{1}_{k}\mathbf{1}_{k}^T).
\]
From\vspace*{1pt} the argument in the proof of Lemma \ref{mu},
$\popl$ has the same nonzero eigenvalues as
$(Z^TZ)^{1/2}B_L(Z^TZ)^{1/2} \in R^{k \times k}$. Let\vspace*{1pt} $\lambda
_1,\ldots, \lambda_k$ be the eigenvalues of $(Z^TZ)^{1/2}B_L(Z^TZ)^{1/2}=
(s^{1/2}I_k)B_L (s^{1/2}I_k) = sB_L$. Notice that $\mathbf{1}_k$ is an
eigenvector with eigenvalue $1$:
\[
sB_L\mathbf{1}_k = \frac{s}{nr + sp}( pI_k + r\mathbf{1}_k\mathbf
{1}_k^T)\mathbf{1}_k = \frac
{s(p+ kr)}{nr + sp}\mathbf{1}_k = \mathbf{1}_k.
\]
Let $\lambda_1 = 1$. Define
\[
\mathcal{U} = \{u \dvtx\|u\|_2 = 1, u^T\mathbf{1} = 0\}.
\]
Notice that for all $u \in\mathcal{U}$,
%
%
\begin{equation} \label{Ueig}
sB_Lu = \frac{s}{nr + sp} ( pI_k + r\mathbf{1}_k\mathbf{1}_k^T) u =
\frac
{sp}{nr + sp} u.
\end{equation}
Equation (\ref{Ueig}) implies that for $i > 1$,
\[
\lambda_i = \frac{sp}{nr + sp}.
\]
This is also true for $i=k$.
\[
\lambda_k = \frac{sp}{nr + sp} = \frac{sp}{nr + sp} = \frac
{1}{k(r/p) +
1}.
\]
This is the smallest nonzero eigenvalue of $\popl$.
\end{appendix}

\section*{Acknowledgments}
The authors are grateful to Michael Mahoney and Benjamin Olding for
their stimulating discussions. Also, thank you to Jinzhu Jia and Reza
Khodabin for your helpful comments and suggestions on this paper.


%

%
\printaddresses

\end{document}